\documentclass[11pt]{article} 

\usepackage[utf8]{inputenc} 
\usepackage{amsmath}
\usepackage{amssymb}
\usepackage{amsthm}
\usepackage{algorithmic}
\usepackage{algorithm}
\usepackage{wrapfig}
\usepackage{subfigure}
\usepackage{afterpage}
\usepackage{tikz}
\usetikzlibrary{matrix,chains,positioning,decorations.pathreplacing,arrows}

\usepackage{cite}

\usepackage{geometry} 
\usepackage{graphicx} 
\usepackage{caption}

\usepackage{booktabs} 
\usepackage{array} 
\usepackage{paralist} 
\usepackage{verbatim} 
\usepackage{subfig} 

\usepackage{fancyhdr} 
\usepackage{sectsty}

\usepackage[nottoc,notlof,notlot]{tocbibind} 
\usepackage[titles,subfigure]{tocloft} 

\title{The Self-Simplifying Machine: \\Exploiting the Structure of Piecewise Linear Neural Networks to Create Interpretable Models}
\author{William Knauth \footnote{wknauth@andrew.cmu.edu (CMU), william@knauth.org (personal)} \\ Carnegie Mellon University \footnote{Research conducted during internship with Wells Fargo Corporate Model Risk}  \footnote{Special acknowledgement to Agus Sudjianto and Rahul Singh of Wells Fargo Corporate Model Risk, who both made significant contributions to this work, and to Wells Fargo Corporate Model Risk in partnership with IBM for providing the computing platform to conduct this research}}
\date{} 

\begin{document}
\maketitle

\begin{abstract}

Today, it is more important than ever before for users to have trust in the models they use. As Machine Learning models fall under increased regulatory scrutiny and begin to see more applications in high-stakes situations, it becomes critical to explain our models. Piecewise Linear Neural Networks (PLNN) with the ReLU activation function have quickly become extremely popular models due to many appealing properties; however, they still present many challenges in the areas of robustness and interpretation. To this end, we introduce novel methodology toward simplification and increased interpretability of Piecewise Linear Neural Networks for classification tasks. Our methods include the use of a trained, deep network to produce a well-performing, single-hidden-layer network without further stochastic training, in addition to an algorithm to reduce flat networks to a smaller, more interpretable size with minimal loss in performance. On these methods, we conduct preliminary studies of model performance, as well as a case study on Wells Fargo's Home Lending dataset, together with visual model interpretation .
\end{abstract}

\section{Introduction}

A Piecewise Linear Neural Network is, by nature, a piecewise linear function. It is well established in the lieterature that a PLNN partitions the input space into regions according to whether each neuron is “on” or “off,”\footnote{This may be verified easily by defining a relation, $x_1 \sim x_2 \iff x_1$ and $x_2$ impose the same configuration, then checking that it is an equivalence relation} and within each region, a linear equation is applied \cite{unwrapping} \cite{openbox}. This structure works well, as any function with some continuous properties can be approximated with a piecewise linear function, provided that the input space is partitioned well. Beyond that, piecewise linear functions are tractable to us for interpretation \cite{unwrapping}. Because of these properties, we work with PLNNs and use the information provided by the linear functions to inform a new transformation of the inputs.

Many attempts have been made to explain what a Neural Network does, but for our purposes it is simple: it learns a nonlinear transformation of the inputs such that the transformed inputs are fit well with a linear model—the simplest model we have. This is similar to the idea of the “kernel trick” in support vector machines, where we implicitly transform the inputs into a high-dimensional feature space, in which the classes are linearly separable. We exploit this feature by constructing a transformation using the linear equations from a large network, which according to our tests, is empirically shown to produce results comparable to the original network without any additional hidden layer training. Using this, we create a single-layer Neural Network, referred to here as “flattened.” The advantages of single-layer networks are that the input transformation is easy to understand and that they may be easily reduced into smaller single-layer networks for improved interpretability and possible resistance to overfitting. 

\subsection{Main Contributions}

This paper seeks to introduce a few main concepts toward improved understanding and interpretability of ReLU Neural Networks. Among them include:

\begin{itemize}

\item A closed formulation of the key features of a PLNN, with special attention given to the properties of networks with a single hidden layer

\item Two main theorems concerning the geometry of linear region boundaries and decision boundaries of PLNN models--first, considering the region boundaries of a single-hidden-layer network, and second, considering the decision boundaries of an arbitrary PLNN

\item A method, called ``flattening,'' utilizing the intrinsic information of a PLNN to design a new single-hidden-layer network, which exhibits similar performance to the original model and facilitates expedient simplification

\item Study of a particular case of network pruning for single-hidden-layer networks, in which we fix the hidden layer and employ the simple process of L2-penalized regression to reduce the size of the network

\item Case study of these processes in action, applied to the real-world dataset on home lending provided by Wells Fargo with visual model interpretation, with discussion of comparative performance and improvements in exact interpretability provided by small, single-hidden-layer networks

\end{itemize}

\subsection{Definitions}

For the purposes of our discussion, we will now introduce and set precise definitions for some key notation and some mathematical background.
\begin{itemize}
\item $N$: number of hidden-layer neurons in a neural network
\item $n$: the number of hidden layers in a network
\item $X$: matrix of inputs 
\item $Y$: vector of class labels for each instance (for binary classification, 0 or 1)
\item $W_i$: the weight matrix associated with layer $i$
\item $B_i$: the bias vector associated with layer $i$
\item Rectifier: the piecewise linear function $\text{ReLU}: \mathbb{R} \to \mathbb{R} \text{ s.t. } x \mapsto \text{ max}(0, x)$. We extend this definition to apply it to vectors as an element-wise ReLU.
\item $\sigma$: the sigmoid activation function $\sigma : \mathbb{R} \to [0, 1] \text{ s.t. } x \mapsto \frac{1}{1+e^{-x}} $
	\item Activation state: whether a neuron’s ReLU returns 0 (off) or $x$ (on), sometimes denoted as 0 or 1, respectively
	\item Configuration: an ordered list of activation states corresponding to each neuron (sometimes referred to in other works as “activation pattern”)
	\item Piecewise Linear Neural Network (PLNN): the function $$F: \mathbb{R} ^m \to [0, 1] \text{ such that }X \mapsto \sigma(W_n(\text{ReLU} (W_{n-1} (\text{ReLU} ( ... (W_1 X + B_1) ... )) + B_{n-1})) + B_n)$$
	\item Linear Equation: the internal part of a linear classification model, $\sigma^{-1}(F_C(X)) = \hat{W}X + \hat{B}$
	\item Partition: set of $A_1, ... , A_t \subseteq \Omega$ such that $\cup_i A_i = \Omega$ and $A_i \cap A_j = \emptyset \, \forall \, i, j$
	\item Region: one such subset $A_i \subseteq \Omega$
	\item Throughout the paper, we will use $$\prod_{i \in [n]}^{*} a_i := \prod_{i=1}^n a_{n+1-i} $$  to denote a left-multiplication product, as opposed to the traditional right-multiplication product. This is to simplify the somewhat cumbersome notation that arises due to noncommutativity in matrix multiplication.

\end{itemize}


%
%
%
%
%
%
%
%
%
%

\begin{figure}[h]
$$\text{\textbf{Configuration in a Toy Model:}}$$\\
$$\text{Hidden layer 1: } a_1 = \text{ReLU}\left(\begin{bmatrix}
3 & 2\\
-1 & 1\\
1 & 0
\end{bmatrix} x + \begin{bmatrix}
2\\
-1\\
-1
\end{bmatrix}\right), \text{ Pre-activations: } 
\begin{bmatrix} 10\\-2\\1 \end{bmatrix}, \text{ Configuration: } [1, 0 , 1]$$

$$\text{Hidden layer 2: } a_2 = \text{ReLU}\left(\begin{bmatrix}
2&1&-5\\
0&7&-4
\end{bmatrix} a_1 + \begin{bmatrix}
-2\\
1
\end{bmatrix}\right), \text{ Pre-activations: } 
\begin{bmatrix} 13\\-3 \end{bmatrix}, \text{ Configuration: } [1, 0]$$

$$\text{Output Layer: } a_3 = \sigma \left(\begin{bmatrix}
1&-4
\end{bmatrix} a_2 + \begin{bmatrix}
-5
\end{bmatrix}\right), \text{ Pre-activations: } 
\begin{bmatrix} 8 \end{bmatrix}, \text{ Class prediction: } 1$$
$$\text{\textbf{Total configuration: [1, 0, 1, 1, 0]}}$$

\centering
    \begin{tikzpicture}[shorten >=1pt,->, draw=black!50,
        node distance = 9mm and 30mm,
          start chain = going below,
every pin edge/.style = {<-,shorten <=1pt},
        neuron/.style = {circle, draw=black, fill=#1,   
                         minimum size=17pt, inner sep=8pt,
                         on chain},
         annot/.style = {text width=4em, align=center}
                        ]
\node[neuron=green!50,
	pin=180:] (I-1) {2};
\node[neuron=green!50,
	pin=180:] (I-2) {1};

\node[neuron=blue!50,
	above right=9mm and 30mm of I-1.center] (H1-1) {10 (on)};
\node[neuron=blue!50,
	below=of H1-1] (H1-2) {0 (off)};
\node[neuron=blue!50,
	below=of H1-2] (H1-3) {1 (on)};
	
\node[neuron=blue!50,
	below right=9mm and 30mm of H1-1.center] (H2-1) {13 (on)};
\node[neuron=blue!50,
	below=of H2-1] (H2-2) {0 (off)};
	
\node[neuron=red!50,
	below right=9mm and 30mm of H2-1.center] (O-1) {1};
	
	\foreach \i in {1,2}
        	\foreach \j in {1,...,3}
{
    \path (I-\i) edge (H1-\j);
    \path (H1-\j) edge (H2-\i);
}
\path (H2-1) edge (O-1);
\path (H2-2) edge (O-1);

    \end{tikzpicture}
\caption{Demonstration of configurations with a toy model PLNN}
\end{figure}
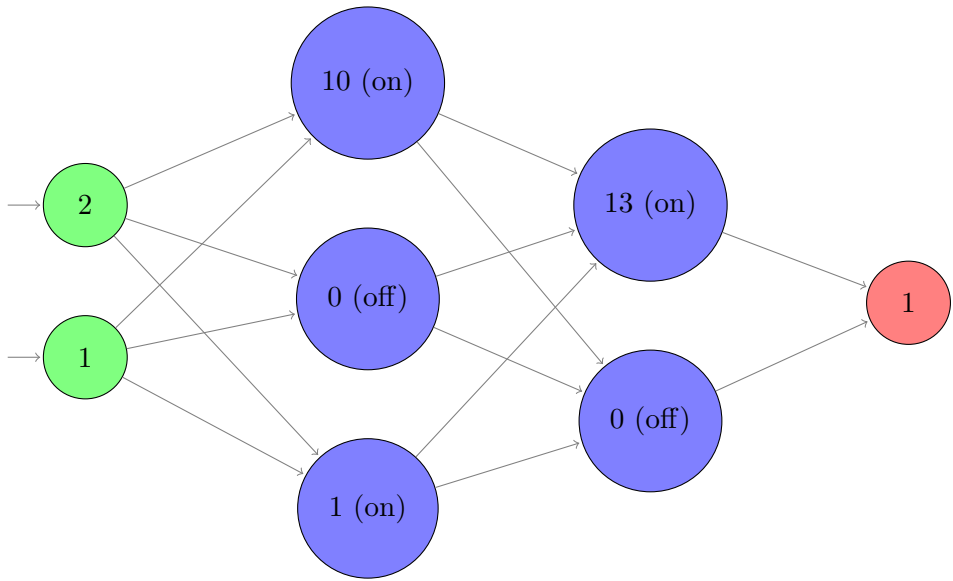

 \subsection{A Network's Configuration}

As discussed earlier, a PLNN partitions the inputs space based on the activation of each neuron. The regions of the input space are defined by a set of linear inequalities corresponding to a particular neuron’s zero activation hyperplane \cite{unwrapping} \cite{openbox}. In a single-layer network, this is easy to understand, so a larger, flattened, single-hidden-layer network itself will also be more interpretable than a deep network. There is one constant set of zero-activation hyperplanes, and each region is defined exactly by which side of each hyperplane the instance is on. We show this in the following theorem, which shows constructively both the partition and the resulting piecewise linear model. A weaker but more general form of this result is presented in \cite{openbox}.\\
\\
\textbf{Theorem 1}:\\
\\
Given a single-hidden-layer network $F$ with $N$ neurons, there exists a constant set $H$ of at most $N$ hyperplanes such that the linear regions of $F$ are convex polytopes defined by a unique set of at most $N$ linear inequalities generated by $H$.
\begin{proof} 
Given $F:\mathbb{R}^m \to [0,1]$ s.t. $X \mapsto \sigma (W_2 (\text{ReLU} (W_1 X+B_1 ))+B_2)$ and a configuration $C$, consider the function $F_C:\mathbb{R}^m \to [0,1]$ s.t. $X \mapsto \sigma ( W_2^C (W_1 X+B_1 )+B_2)$, where $W_2^C$ is defined such that $(W_2^C)_i=C_i (W_2 )_i$, and $(M)_i$ denotes the $i$th column of $M$. 

Note that in feedforward computation, if $F$ has configuration $C$ when $x$ is passed as the input, then all neurons with activation state 0 will feed forward a 0. This is the same as the corresponding column of $W_2$ consisting of only zeros, since $(W_2 )_i \cdot 0=0 \cdot (W_1 x+B)_i=0$, hence our choice of $W_2^C$ in $F_C$. Therefore, for all $x$ yielding configuration $C$, we have that $F(x)=F_C (x)$. Further, $F_C (x)=\sigma (W_2^C W_1 x+(W_2^C B_1+B_2 ))$, so all such $x$ share the same linear equation.

To establish that such linear regions are convex polytopes, the Hyperplane Separation Theorem due to Minkowski \cite{convex} gives us that it is sufficient to define each region by a set of linear inequalities. 

We now define a constant set $H$ of at most $N$ hyperplanes as $$H=\{S_i \subseteq \mathbb{R}^m : x \in S_i \iff (W_1 x+B_1 )_i=0,i \in [N]\}$$. 
Let $\Omega$ denote the space of all linear inequalities. Further, we define $\alpha:{0,1}^N \to {<,\geq}^N$ as the natural bijective correspondence between lists of 0s and 1s and lists of $< and \geq$, with 0 corresponding to $<$ and 1 corresponding to $\geq$. We define the generated inequalities of $H$ to be the function $$\beta_H:{<,\geq}^N\to \Omega \text{ such that }$$ $$(\sim_1,\sim_2,…,\sim_N )\mapsto \{ (W_1 x+B_1 )_1  \sim_1  0,(W_1 x+B_1 )_2  \sim_2  0,…,(W_1 x+B_1 )_N  \sim_N  0 \}.$$\\ Note here that $\sim_i$ denotes either $<$ or $\geq$. We now define the function $$\gamma_H :{0,1}^N\to \Omega \text{ such that }\gamma_H = \beta_H \circ \alpha.$$ To complete the proof, we now must check that the subset $A_C \subseteq \mathbb{R}^m$, defined as all $x\in \mathbb{R}^m: x \in A_C$ implies $F$ has configuration $C$, is the same as the subset $B_C$ defined as all $x \in \mathbb{R}^m$ such that $x$ satisfies all linear inequalities in $\gamma_H(C)$. 

We first verify that $A_C\subseteq B_C$. Let $x\in A_C$ and $i \in [N]$. Then given $C$, we note that $C$ is an ordered list of activation states corresponding to each neuron. For the $i$th neuron, we note that its activation state is defined by whether ReLU($(W_1 x+B_1 )_i$) returns 0 or $(W_1 x+B_1 )_i$. This is defined by whether $(W_1 x+B_1 )_i$ is less than $0$ or greater than or equal to zero. So $(W_1 x+B_1 )_i  \sim_i  0$ must be true. Therefore $x\in B_C$. 

We now verify that $B_C\subseteq A_C$. Let $x \in B_C$. So $x$ satisfies all inequalities in $\gamma_H ( C)$. So, given the $i$th neuron, we know that $x$ satisfies $(W_1 x+B_1 )_i   \sim_i   0$, and therefore, the neuron will have the correct activation state. So $x \in A_C$, and thus $A_C=B_C$. Thus the existence is established. One may easily verify that $\gamma_H$ is injective, and this establishes the uniqueness, as no two configurations may map to the same set of inequalities. Therefore, our proof is complete. 

\end{proof}


We have now demonstrated for a single-hidden-layer network, in closed form, the partition of the input space such that within each region we have a local linear model. With a deep network, the partition may be more complicated and difficult to understand, as a neuron’s zero activation hyperplane changes based on the activation states of neurons in previous layers. Due to this, converting a network to a single layer provides us with a concise, exact local interpretation: we can check a single, relatively small list of inequalities and subsequently choose from a list of linear models to apply.

\subsection{General Formulation}

In more generality, the model becomes more complicated. We now present, in closed form, exact piecewise linear interpretation of an arbitrary PLNN; further details can be found in \cite{unwrapping}. We may imagine the space of all possible inputs that would share a particular configuration when passing through an arbitrary PLNN, forming again a region of the partitioned input space. We consider which neurons are “off” in activation state. Suppose, for sake of illustration, that neuron $i$ (only) is off in layer $l$. Then we have that within that region, if we define $\tilde{W_l}$ as $W_l$ with the $i$th column set to zero, $$F_C (X)=\sigma(W_n (…(\tilde{W_l }(W_{l-1} (…(W_1 X+B_1 )…)+B_{l-1} )+B_l )…)+B_n).$$ 

This is because of the fact that if that neuron’s ReLU activation is always zero, it is equivalent to all weights associated with that neuron being set to zero. 

We may then expand this idea to any subset of a network’s neurons. For any given configuration, we may set all necessary columns of weight matrices to zero and thus eliminate the ReLU nonlinearities. We will denote the new weight matrices as $W_l^C$, and note that $W_1^C=W_1  \, \forall \, C\in \{0,1\}^N$. To do this, we first isolate the starting and finishing indices, $i$ and $j$,  within the configuration $C$ for layer $l$. We then define the matrices $\hat{C}_{[i:j]}$  by casting the slice $C[i:j]$ into a diagonal matrix. This gives us a matrix similar to an identity matrix, only with some diagonal elements set to zero. Finally, we obtain $W_l^C$ by $$W_l^C := W_l \hat{C}_{[i, j]}.$$


Removing all ReLU nonlinearities in this manner gives us that within that region, $F_C (X)=\sigma(\hat{W}X+\hat{B})$ for some $\hat{W}, \hat{B}$. We refer to the internal part, $\sigma^{-1} (F_C (X))=\hat{W}X+ \hat{B}$ as the \textit{Linear Equation}. Each region has a linear equation, and we refer to the set of the linear equations over all regions in a network as the network’s linear equations. Calculating such $\hat{W}, \hat{B}$ is simple. We first apply the configuration to the weight matrices as described above. It then follows that: $$\sigma^{-1} (F_C (X))=\left( \prod_{l\in [n]}^*W_l^C \right)X+\left[\sum_{i\in[n-1]} \left(\prod_{l>i}^* W_l^C \right) B_i \right]+B_n.$$

Further, consider the possibility that in a configuration, all neurons in a given layer are “off.” We imagine setting the columns of the following weight matrix to zero and quickly realize that this leads to $$F_C (X)=\sigma(\bold{0}X+B)=\sigma(B) $$ for some bias $B$, or in other words, the output has no dependence on the input. We call these regions \textit{trivial regions}, and no decision boundaries may pass through them—all instances within a trivial region will receive the same predicted class label.

Next, we note that any nontrivial region carries with it a decision boundary due to its linear equation\footnote{Note it is possible that such a decision boundary may not pass through the sample data, or even the geometric region itself, so there may exist many nontrivial, yet still single-class regions}. Such a decision boundary is easy to calculate. With sigmoid activation, we solve quickly that:
$$F_C (X)=0.5 \implies \sigma^{-1} (F_C (X))= \hat{W}X+ \hat{B}=0.$$

We now imagine how to find all linear inequalities that define a region for a general PLNN. The process is straightforward: in each configuration, each neuron in the network will come with a corresponding linear inequality generated by its zero-activation hyperplane. This is due to the fact that crossing the neuron’s zero-activation hyperplane will move you to a different configuration and thus a different region. Note that in contrast with a single-hidden-layer network, a deep network’s zero-activation hyperplanes change based on the configuration, due to ReLU nonlinearities in hidden layers before the neuron in question. Given a configuration $C$ and the $i$th neuron in layer $L$, we calculate its zero-activation hyperplane much in the same way we would calculate a regional decision boundary:

$$\left(\left( \prod_{l \in [L]}^*W_l^C \right) X + \left[ \sum_{j \in [L-1]} \left( \prod_{L \geq l > j}^* W_l^C \right) B_j \right] + B_L \right) _i = 0.$$

Given these closed form solutions for the partitioned space and the linear equations that govern our model, we are now equipped to proceed.

	The goal of our work is to use the intrinsic information of a PLNN in order to reduce the model to the smallest size possible. Such a small model has significant advantages, including interpretability and faster real-time computation. To illustrate the interpretability advantage, we show interpretation plots of two different networks on the similar synthetic data—one with 5 layers of 4 neurons, and one with a single layer of 2 neurons.

\begin{figure}[h]
\centering
\includegraphics{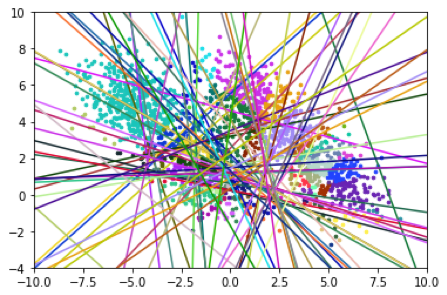}
\includegraphics{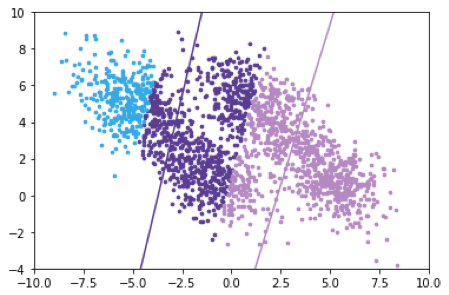}
\caption{Region and decision boundary visualization: on the left is a network with 5 layers of 4 neurons, and on the right is a network with one layer of 2 neurons. The points are instances, colored based on which region (configuration) they are in, and the lines are regional decision boundaries, paired with their region color-wise. For a description of these data, see Section 3. The larger network has many more regions and is confusing to interpret. On the other hand, it is very clear to see how the small network makes its decision.}
\end{figure}
	
	
\section{Related Works}

Research has shown that large, deep networks are more reliable to train \cite{lottery}. Many attempts have been made to bypass this apparent tradeoff between large and small models, and \cite{lottery} offers one possible solution. It attempts to find smaller subnetworks of a well-trained large network that perform similarly to the larger network. The “Lottery Ticket Hypothesis” \cite{lottery} says that within a large network, there exists a small subnetwork with equal performance, despite the fact that re-initializing and re-training the smaller subnetwork leads to worse performance. This work advances our cause in two ways—first, it showed that larger models are indeed more reliable in training (even if a small network is sufficient to solve the problem), and second, it showed that a large model can be reduced effectively. The Lottery Ticket paper further showed that their search for a subnetwork often reached similar results to those of traditional pruning.

	In our approach, we do not attempt to directly find subnetworks of a large network; instead, we extract the underlying structure of a given PLNN. In their paper, Chu et al devise a method for analyzing the model structure of a PLNN \cite{openbox}. They showed that the network divides the input space into regions that are convex polytopes, meaning that they are defined by a set of linear inequalities. Further, they showed that within each convex polytope, the model acts as a linear classifier with a closed form solution. With this information, they introduce the \textit{OpenBox} algorithm for interpretation, which iterates through the test data, detects configurations, and compiles a list of all active regions and the linear classifiers used within each region. We utilize this information in the process of network flattening described in Section 3.

	The Extreme Learning Machine established that it is possible to conduct training of a neural network model only on the output layer \cite{extreme}. In the Extreme Learning Machine, the hidden layers are all initialized randomly and are not trained, with simple regression performed to determine the output layer parameters. This shows that in practice, a transformation into a feature space does not necessarily need to be fine-tuned for a model to perform acceptably. We apply this idea to network flattening, using the piecewise linear equations found by deep networks. This approach provides a construction of the transformation that is empirically shown to be reliable.
	
	In such a transformation, it is natural to wonder what exactly a ReLU nonlinearity does to help linear separability. Previous studies show the computable functions by neural networks and establishes a geometric interpretation for what ReLU nonlinearities do to transformed data \cite{onTheNumber}. The nonlinearities serve to “fold” the input space so that a single hyperplane can separate the data. Successive hidden layers serve to fold the feature spaces provided by previous layers’ transformations, adding functional complexity, which would suggest an advantage to deep learning. While this interpretation is useful, we explore the possibilities of what single-layer networks can do.
	
	One answer to this lies in the well-known Universal Approximation Theorem, which shows that for any nonconstant, bounded, and continuous activation function, the set of functions computable by a single-hidden-layer neural network is dense in the set of continuous functions on a compact subset of $\mathbb{R}^m$ \cite{universalapprox}. To be precise, for any valid activation function $\phi:\mathbb{R}\to \mathbb{R}$, for any $\epsilon>0$, and for any continuous function $f: K \subset \mathbb{R}^m \to \mathbb{R}$, where $K$ is compact, there exists a single layer network $F:K\subset \mathbb{R}^m \to \mathbb{R}$ such that $|F(x)-f(x)|< \epsilon  \, \forall \, x \in K$, or in other words, $F$ is a pointwise approximation of $f$ to within $\epsilon$. It was further shown that this property is obtained if and only if $\phi$ is nonpolynomial \cite{nonpolynomial} \cite{pinkus}. In our case, the ReLU activation function is indeed nonpolynomial, so the universal approximation theorem applies to a single-layer PLNN.
	
	Finally, note that \cite{unwrapping} contains a great deal of similar machinery to this paper, as the research contributing to both papers was conducted largely together. For instance, \cite{unwrapping} makes refined use of the flattening algorithm (Algorithm 1), presented in its original form here in Section 3. Further, the Python package \textbf{Aletheia}$^\mathbf{\copyright}$, introduced in \cite{unwrapping}, contains much of the code required to carry out our techniques.


\section{Methodology}

\subsection{Network Flattening}

Due to the work of the Extreme Learning Machine \cite{extreme} and the Universal Approximation Theorem \cite{universalapprox} \cite{nonpolynomial} \cite{pinkus}, it is reasonable to attempt construction of a single layer network without all of the difficulties that come with training a large model. We use the linear equations (obtained through the methods described in Section 1), applied to a deep network, to create a “reservoir.” We then represent it with a flat, single-hidden-layer arrangement, and then solve only a logistic regression problem to optimize the output layer. We see empirically in Section 4 that the flattened networks obtained from well-trained deep networks do not appear to have an accuracy disadvantage in initial testing. Having only a single layer does not limit what functions can be computed, and the reservoir transformation we compute separates the data quite well.

Below, we present the main algorithm of our methods. Algorithm 1, flatten, determines the linear equations used by the network in its different regions, and then transforms the data into a space in which these equations are the new coordinates. We then take the element-wise ReLU activation of each new data point. Empirically, we see that the data can be separated comparatively well by a linear model in this new space. This transformation and linear model can then be represented equivalently by a single-layer neural network. The main intuition is that given enough information on where an instance stands relative to many trained decision boundaries, logistic regression suffices to learn the entire model.

We now ask the reader to recall Theorem 1, in which we show that configurations are shared by regions defined by the hyperplanes of the hidden layer transformation. In this case, the hidden layer transformation contains the decisions of each linear model. Therefore, the decision boundaries of the original model become the partition boundaries of the flattened network. The resulting partition will divide the input space into regions based on which subset of $H$, the set of $ N$ hyperplanes an instance is “above,” or in this case, which of the original network’s linear equations return a positive value. This gives us regions in which class predictions should be very similar, or in other words, almost all of the work is done by the hidden layer. Logistic regression then serves to fine-tune the model in order for it to perform well.

\begin{algorithm}
\caption{flatten}
\begin{algorithmic}
 \STATE inputs: PLNN $F$; training data $X, Y$
 \STATE extract active configurations, linear equations, and basic architecture from network $F$
 
 \STATE $k \gets 0$
 \STATE $M, V \gets$ empty matrix, empty vector
 
 \FORALL{active configurations} \STATE{
 $w, b \gets$ linear equation weight, bias}
 \IF{$w \neq \mathbf{0}$} 
 
 \STATE{$k \gets k+1$}
 \STATE{add row to $M$, fill with $w$}
 \STATE{add row to $V$, fill with $b$}
 
 \ENDIF
 
 \ENDFOR
 
 \IF{$k = 0$} \RETURN{original network}
 \ENDIF
 
 \STATE $X_k \gets$ ReLU$(MX + V)$
 \STATE fit (regularized) Logistic Regression model $F*$ to $X_k, Y$
 \STATE $W, B \gets $ weight, bias from $F*$
 \RETURN Model $F_{\text{flat}} = \sigma(W(\text{ReLU}(MX + V)) + B)$

 \end{algorithmic}
 \end{algorithm}

A key point to note is that this algorithm involves no additional hidden-layer training of a new network. The additional training is done for the output layer only using logistic regression, which is a convex optimization problem and therefore much easier than the training of a network. In short, we can efficiently use well-performing deep networks to produce well-performing single-hidden-layer networks without additional hidden-layer training. 

The basis of this algorithm is the idea of transforming the data according to the linear equations from the previous network. Adding the ReLU nonlinearity serves two main purposes. First, it makes “folds” in the data so that the data are better represented by a single linear equation \cite{onTheNumber}. Second, it mimics the behavior of the data in the first layer of the network. This allows us to perform Logistic Regression to find the output layer weight and bias, which yields a global minimum. This is a reliable process as compared to the nonconvex problem posed by optimizing a neural network. 


\begin{wrapfigure}{l}{0.4\textwidth}
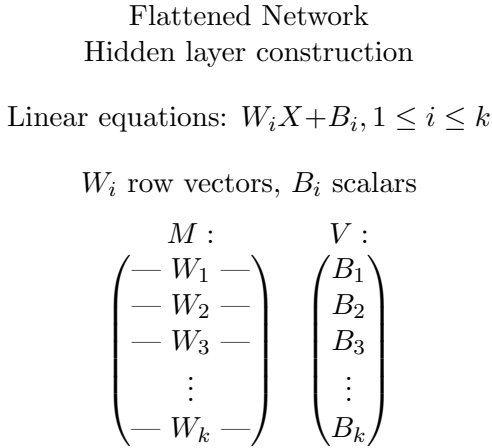

	\centering
	Flattened Network \\Hidden layer construction
	$$\text{Linear equations: } W_i X + B_i, 1 \leq i \leq k$$
	$$W_i \text{ row vectors, }B_i\text{ scalars}$$
	$$\stackrel{\stackrel{\mbox{$M:$}}{\mbox{ }}}{%
	\begin{pmatrix}
	\text{--- } W_1 \text{ ---} \\
	\text{--- } W_2 \text{ ---} \\
	\text{--- } W_3 \text{ ---} \\
	\vdots \\
	\text{--- } W_k \text{ ---} 
	\end{pmatrix}}\, \, \, \, \, \, \stackrel{\stackrel{\mbox{$V:$}}{\mbox{ }}}{%
	\begin{pmatrix}
	B_1\\
	B_2\\
	B_3\\
	\vdots \\
	B_k
	\end{pmatrix}}$$
	\caption{Construction of $M$, $V$ from Algorithm 1}
\end{wrapfigure}

The $k = 0$ case may raise questions about the robustness of the algorithm; however, it is used extremely rarely according to our testing, especially in practice, when we would be flattening only large networks. This is because $k = 0$ would mean that the network divides the data only into trivial regions—ones in which there exists a layer such that all neurons in the layer are “off.” The probability of this happening in practice with a large network is very small.

\subsection{Pruning a Flat Network}

In addition to simply creating a flat network, we may reduce, or “prune,” this flattened network further, through an iterative process of deleting unnecessary neurons. In order to do this effectively, it is important to understand the effects of a single neuron on the network.

We again ask the reader to recall Theorem 1. Through our construction, it is clear to see that the region boundaries in our partition are defined by the zero-activation hyperplanes of each neuron. Therefore, the effect of a single neuron is in fact this region boundary, along with the change in local linear model associated with crossing the region boundary.
To understand the effect of a single neuron is to understand the relationship between the local linear models applied to adjacent regions. Our research finds that regional decision boundaries must intersect when crossing region boundaries (which correspond to the effect of a single neuron) and the decision boundaries must intersect at an angle determined by the relative magnitude of coefficients.\\
\\
\textbf{Theorem 2}:\\
\\
Decision boundaries corresponding to adjacent nontrivial regions of a PLNN intersect at the region boundary.
\begin{proof}
We begin by considering the region boundary determined by neuron $i$ in layer $l$. Each region shares a configuration--we will denote these $C_1$ and $C_2$. For all $k \neq l$, $W_k^{C_1} = W_k^{C_2}$, so we will just denote these as $W_k^C$. The regional decision boundaries are:

$$W_n^C (…(W_l^{C_1} (a_l )+b_l )+...)+b_n=0,$$
$$W_n^C (…(W_l^{C_2} (a_l )+b_l )+...)+b_n=0,$$ 
where $a_l$ is the vector of neuron outputs at layer $l$. We may solve these equations for the intersection of decision boundaries to obtain that 
$$\left( \prod_{k>l}^* W_k^C \right)(W_l^{C_1}- W_l^{C_2}) a_l=0.$$

We observe that $(\prod_{k>l}^* W_k^C )$ is a row vector and that $(W_l^{C_1}- W_l ^{C_2} )$ is a matrix with nonzero values restricted only to column $i$. The resulting product $\left( \prod_{k>l}^* W_k^C \right)(W_l^{C_1}- W_l^{C_2})$ will thus be a row vector with nonzero values restricted only to position $i$. 

Defining $v= \left( \prod_{k>l}^* W_k^C \right)(W_l^{C_1}- W_l^{C_2})$, we will therefore view the equation $\left( \prod_{k>l}^* W_k^C \right)(W_l^{C_1}- W_l^{C_2}) a_l=0$ as a standard inner product, $\langle v,a_l \rangle= \sum v^k a_l^k  =v^i a_l^i=0$. Because the real numbers are an integral domain, we may now conclude that either $v^i=0$ or $a_l^i=0$. If $v^i=0$, then we recognize that in both configurations, the network’s output is independent of the neuron. So, the two regions employ the same linear equation and thus decision boundary, or in other words, the decision boundaries’ intersection is the entirety of each. If $a_l^i= 0$, then we have that the decision boundaries intersect precisely at the region boundary. So in either case, the decision boundaries must intersect at the region boundary. 

\end{proof}

Since we have identified the restricted nature of boundary intersections, we may further investigate by considering how similarly these two regions will perform the classification task. The cosine similarity of boundary normal vectors is an intuitive and useful measure of how alike two regions behave. Such a boundary normal vector is simply the vector $\hat{W}^C=W_2^C W_1$ in a single-layer network. We may compute this cosine similarity easily: 

$$c = \frac{\langle W_2^{C_1}W_1, W_2^{C_2}W_1 \rangle}{
|W_2^{C_1}W_1| |W_2^{C_2}W_1|} = \frac{
W_1^T (W_2^{C_1})^T W_2^{C_2} W_1}{\sqrt{
W_1^T (W_2^{C_1})^T W_2^{C_1} W_1 \cdot W_1^T (W_2^{C_2})^T W_2^{C_2} W_1}}.$$

We observe that this quantity is clearly dependent on the relative magnitude of the output layer weight associated with this neuron\footnote{The details of this calculation can be messy and are not overly enlightening, and therefore are omitted. However, do note that the key is in the numerator, where we multiply $(W_2^{C_1})^T W_2^{C_2}$. The two matrices are the same, save for one column of zeros, and the extra zeros play the central role.}. One may verify easily that 

$$ \lim_{(W_2)_i \to 0} c = 1,$$
and that this is indeed a global maximum. This, along with the obvious continuity, means there exists a neighborhood of 0 in which the cosine similarity will be arbitrarily close to its maximum. In other words, if the weight associated with a neuron is small, the adjacent regions separated by the neuron's zero-activation hyperplane will have extremely similar linear equations.

In eliminating a neuron, we are effectively removing some series of nonlinearities from the model by merging some adjacent pairs of regions. Upon re-fitting the output layer, the effect will be close to an “average” of the linear equations for each pair of regions across the region boundary that is removed. So, the larger the cosine similarity, the less significant the nonlinearities we remove, and the more accurate each region will remain to be. In practice, this means that we may prune a neuron with comparatively small weight without drastic change to the model. We use L2 regularization in our logistic regression to encourage some weights to become small, so that this pruning works best. We deem these neurons with the smallest output layer weights to be the least important, so we may delete them, and again use logistic regression only to determine the output layer of the new network. 

Further, we see empirically that the effect of this selection is to maintain diversity in our model. If a significant nonlinearity is captured by the flat network, using the L2 penalty and selecting based on output layer weight will serve to preserve this nonlinearity through the pruning process, rather than eliminating the neuron responsible for it. This is because we prune neurons associated with high cosine similarities, and in turn, insignificant (or even overfitting) nonlinearities. We will discuss this observation in greater detail in Section 4.3.

To this end, we introduce the second algorithm of our methods.

\begin{algorithm}
\caption{pruneFlatNetwork}
\begin{algorithmic}
\STATE inputs: $F = \sigma (W(\text{ReLU}(MX + V)) + B)$, a single-layer PLNN;  training data $X, Y$; desired number of neurons $k$
\STATE extract $W, M, B, V$ from $F$
\STATE calculate criterion $c_i$ for each neuron $n_i$ (implemented here as the output layer weight)
\STATE calculate rank, $R(c_i)$ for each neuron
\STATE $M', V' \gets$ matrix, vector containing exactly one copy of row $i$ of $M, V$ if and only if $R(c_i) \leq k$
\STATE $X_k \gets M'X + V'$
\STATE $F* \gets$ logistic regression model trained on $X_k, Y$
\STATE $W', B' \gets$ weight, bias extracted from $F*$
\RETURN model $F' = \sigma(W'(\text{ReLU}(M'X + V')) + B')$


\end{algorithmic}
\end{algorithm}

Doing this process iteratively, deleting one neuron at a time until dropping below a specified performance threshold, gives a simplified and more interpretable model.  Additionally, this provides an empirical upper bound on the size of flat PLNN necessary to solve a certain problem. 

	We may also attempt to expedite the pruning process by deleting a large number of neurons at a time. While this is possible to do with still acceptable results, it is more reliable to conduct the process one-at-a-time, a backward elimination in step-wise regression. This is because it is possible that many consecutive small decision boundary changes exist, all bent in a similar direction, approximating a broad, curved nonlinearity. Then, deleting all of those corresponding region boundaries may cause significant loss for the model, whereas deleting one-at-a-time can detect when enough of those regions have been merged. In a slightly less heuristic sense, by re-regressing at each step, the ordering of the neurons with the smallest output layer weights may change, so in deleting many neurons at once, we may be deleting neurons that would become more necessary at the steps we skip.

\section{Experiments and Data}

We now present results of our experiments testing the two algorithms. In our experiments, we consider two datasets. First, a synthetic dataset based on random sampling from four bivariate normal distributions, and second, the MNIST dataset with the output classes reduced modulo 2 (binary classification of even vs. odd) \cite{MNIST}. The MNIST dataset takes handwritten digits and places them centered in a 28x28 pixel grayscale image. See below for a visual representation of our synthetic dataset. Our maximum observed accuracies on the MNIST dataset were just above 98\%, while our maximum observed accuracies on the synthetic data were just above 94\%. With all training, we utilized the Adam optimizer, implemented in pytorch, with a learning rate of 0.02 \cite{adam}.

\subsection{Flattening Tests}

In the first experiment, we test the flattening algorithm, with the basic question: does flattening work? Are there overall losses to the performance of the flattened network with respect to the performance of the original? 

To answer this question, we consider the paired distribution of networks with their flattened counterpart. We repeat this for varying network architectures on both synthetic data (5000 data points, split into 3000 training and 2000 testing) and MNIST data modulo 2 \cite{MNIST}. 

\begin{wrapfigure}{l}{0.45\textwidth}
    \centering
    \includegraphics[width=0.45\textwidth]{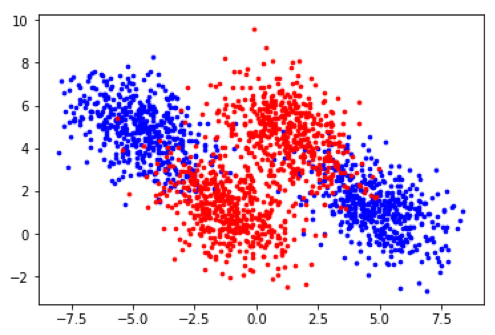}
    \caption{Synthetic binary classification data used in experiments}
\end{wrapfigure}

Our network architectures were as follows: single layer networks with 2, 5, 10, and 50 neurons, two-layer networks with 2, 5, 10, and 30 neurons each, and 5-layer networks with 2, 5, 10, and 20 neurons each. Each network was trained with a batch size of 4. The null hypothesis is that the mean is zero, or in other words, the network and its corresponding flat network are expected to perform the same. 

We conduct a paired $t$-test. For each network architecture, we independently initialize and train 25 networks and for each network, measure the accuracy, flatten the network, measure the size of the flat network, and measure the accuracy of the flat network. If the original network consists only of trivial regions, we discard the data point as flattening is not possible. We present means, sample standard deviations, and $t$-statistics below. Negative mean values and large (significant) $t$-statistics indicate that the flat networks performed worse, and positive mean values and large (significant) $t$-statistics indicate that the flat networks performed better.

\begin{table}
\begin{center}
\begin{tabular}{| l || l | l | l | l |}
\hline
\multicolumn{5}{|c|}{Synthetic Data} \\
\hline
\hline
Network Structure	&Original Mean&	Flattened Mean	&Diff. std. dev.&	$t$-score\\
\hline
\hline
2	&0.85222&	0.850468	&0.082964	&-0.10559\\

5	&0.9326	&0.92982	&0.0005974	&-2.32673\\

10	&0.93532	&0.93646	&0.003315	&1.719396\\

50	&0.935452	&0.93958	&0.00317	&6.511624\\

2x2	&0.860261	&0.863457	&0.09282&	0.165114\\

5x2	&0.932104	&0.93254&	0.00492	&0.443088\\

10x2	&0.93416	&0.93744	&0.002269	&7.227458\\

30x2	&0.9337	&0.93826	&0.004505	&5.060631	\\

2x5	&0.8374	&0.9189	&0.113598	&2.268747	\\

5x5	&0.9039	&0.93358	&0.093897	&1.548522	\\

10x5	&0.93304&	0.93726	&0.006718	&3.140985	\\

20x5	&0.93348	&0.93744	&0.003446	&5.746175	\\
\hline
\hline
\multicolumn{5}{|c|}{MNIST Data} \\
\hline
\hline
Network Structure	&Original Mean&	Flattened Mean	&Diff. std. dev.&	$t$-score\\
\hline
\hline
2&0.890548&	0.877562&	0.023786	&-2.18384\\
5&0.898256	&0.882039	&0.025971	&-2.99463\\
10&0.911472	&0.912108	&0.03442	&0.092387\\
50&0.950184	&0.948976	&0.004212	&-1.43384\\
2x2&0.913031	&0.886554	&0.039792	&-2.3991 \\
5x2&0.912748&	0.897392	&0.023396	&-3.28182\\
10x2	&0.95098	&0.953904	&0.005528	&2.644547\\
30x2&0.968528	&0.972504	&0.0033	&6.023717\\
2x5&0.899667	&0.8943	&0.012587	&-1.47698\\
5x5&0.94466	&0.942416	&0.009295	&-1.20713\\
10x5&0.96752	&0.969132	&0.00264	&3.052702\\
20x5&0.97478	&0.977828	&0.002917	&5.223918\\
\hline

\end{tabular}
\caption{Flattening Experiment Results}
\end{center}
\end{table}
\begin{figure}[h]
\centering
\subfigure[Synthetic Data]
{
\includegraphics[width=0.48\textwidth]{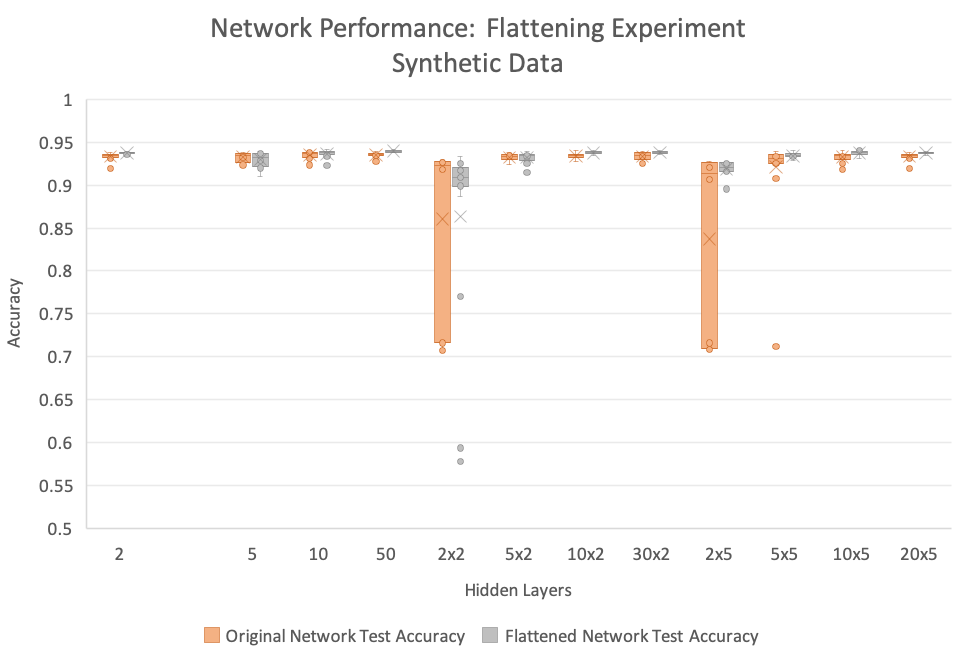}
}
\subfigure[MNIST Data]
{
\centering
\includegraphics[width=0.48\textwidth]{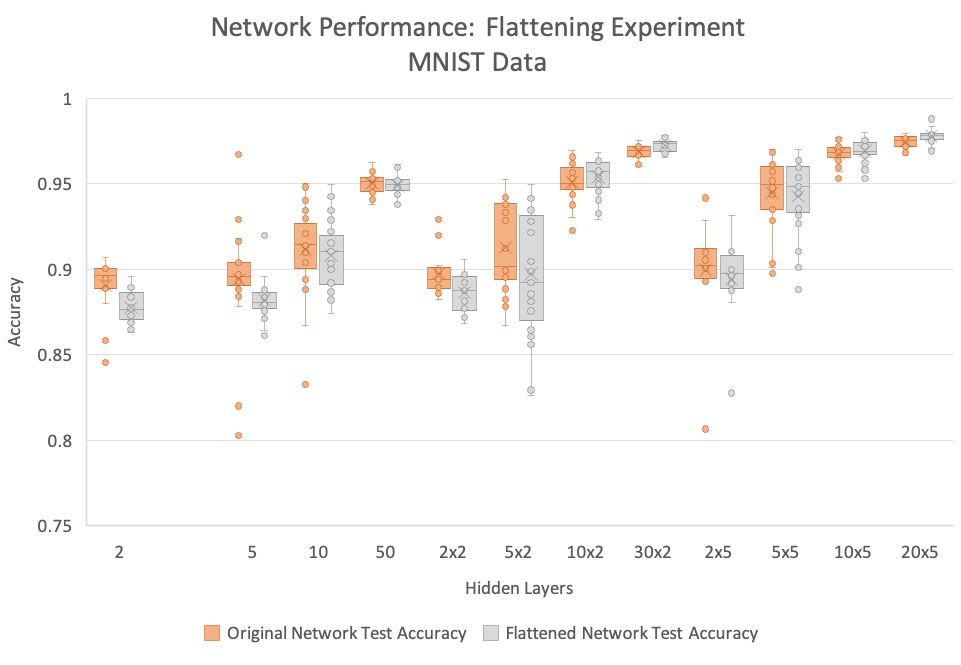}
}
\caption{Flattening experiment results}
\end{figure}

Here we provide an illustration of our experiment results using box-and-whisker plots for each network structure. What this shows us is that in many cases, the variance in performance of flattened networks is very small, and is in most cases smaller than the variance in performance of the original network. In other words, the original network does not need to be perfect in order to yield great results from flattening. What we do see is that flattening tends to perform significantly better on larger networks, but still acceptably on smaller models. In fact, we see that our best performing networks tend to be our largest flattened networks. This reflects the tendency in the above table for significant results in favor of flattened networks on larger architectures.

There are some cases in which we obtain a $t$-score that is very large in favor of the flat networks, which we do not have a full explanation for. One possible reason is that these cases tend to be the ones with very large models, and research shows that larger-dimensional transformations are more likely to leave the data linearly separable \cite{extreme}. In general, we see a trend that flattening becomes more effective as we use larger networks. If the original network is underfitting, it may not carry enough information in its linear equations to be able to separate the data. However, as we increase the size of the network (and thus the dimensionality of the hidden-layer transformation), we see that the flat network becomes very effective.

	Our conclusion is that for sufficiently large networks, the flattening algorithm shows no evidence of systematic performance loss between the original network and the flattened network. While it may be possible with more data collection to establish stronger results, this conclusion is sufficient to support the idea that flattening a network does in fact work. The reason we state “sufficiently large” is that the techniques discussed here are mainly of use for larger-than-necessary network architectures, as smaller networks may be interpreted well as-is. We encourage further testing of this method using many types of data and many different architectures in order to better understand its behavior. 
	
Finally, we note that although these are the only two datasets we have used for structured experimentation, during development we tested the method using various other datasets in the UCI Machine Learning data repository \cite{uci} and from scikit-learn  \cite{scikit-learn}, including HIGGS, Abalone (young vs. old classification), Breast Cancer Prediction, and Taiwan Credit Card Default data. We did not notice a pattern of performance losses upon flattening using these data either. 


\subsection{Pruning Tests}

For our second experiment, we ask the basic question: does pruneFlatNetwork work? To test this, we used the same datasets as the first experiment and applied our pruning process iteratively. We start with an original deep network, flatten it, and prune neurons until the model has only one neuron. For the sake of data entry and the magnitude of the experiment, we pruned half of the neurons at each step until we got under a certain threshold for each dataset. We used 15 as the threshold for MNIST and 20 as the threshold for our synthetic data. However, in practice, it would be advisable to prune one neuron at a time when dealing with a single model, as we have observed this to give better and more consistent results. 

\begin{figure}[h]
\centering
\includegraphics[width=0.8\linewidth]{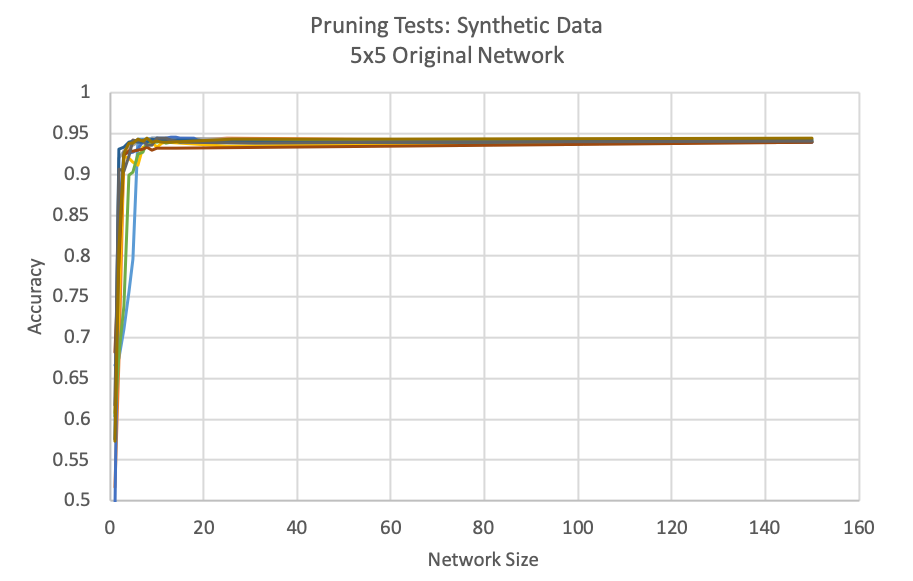}
\caption[width=0.8\linewidth]{Accuracy of 10 pruned networks by size--each color represents an original network, then pruned to 1 neuron with testing accuracy measured at each step}
\label{figure:fig11}
\end{figure}

For each original network architecture, we ran 10 tests of iterative pruning. We did this for three architectures on each dataset. On MNIST, we used 10x2 networks, 30x2 networks, and 20x5 networks. On the Synthetic Data, we used 5x2 networks, 10x2 networks, and 5x5 networks. Note that many of these networks are larger than necessary to carry out this process—we employ large models to demonstrate that our methods can be used on models of any size. Each original network with MNIST was trained using a batch size of 256, and each original network with the synthetic data was trained using a batch size of 32. We record the size of the model at each step and its accuracy, then plot the results.

Although our original models are not flat networks, we record their size as a number larger than all flattened sizes. For synthetic data, the numbers we use are 50 for 5x2, 150 for 10x2, and 150 for 5x5. For MNIST data, the numbers we use are 600 for 10x2, 9000 for 30x2, and 10000 for 20x5. The full results can be found in the Appendix, Figures 6 and 9 through 13. As an example, we include the results from flattening and pruning a 5x5 networks using synthetic data (Figure 6).

We see that in every case, there is a region where we may prune with no loss in performance. In many cases, we are able to prune the network to a very small size before seeing any significant drops in performance. In our developmental testing, we observed that pruning one neuron at a time did lead to more consistent results, so further work includes a more detailed study of pruning in which we prune one neuron at a time\footnote{Such extensive testing will require a computing platform beyond what we were able to use.}.


\subsection{Case Study: Visual Model Interpretation}

To demonstrate the effectiveness of our methods, we now present results of applying our simplification process to a model using real-world data. We use Wells Fargo’s Home Lending dataset, modified for privacy purposes, with binary classification of 0—customer does not default and 1—customer defaults. The dataset has 1 million instances, of which about 1\% are defaulting cases, and 55 predictors. We sampled and split the data into training and testing sets, each with approximately 1/3 of the total instances, using the same split as \cite{slim} \cite{axnn}. We list some of the main predictors and their meanings in the table below, adapted from \cite{slim} \cite{axnn}:

\begin{table}[h!]
\begin{center}
\begin{tabular}{ | l | l | }
\hline
Variable	& Meaning\\
\hline
\hline
fico0	&FICO at prediction time\\
ltv\_fcast	&Loan to value ratio forecasted\\
dlq\_new\_delq0	&Delinquency status: 1 means current, 0 means delinquent\\
unemprt	&Unemployment rate\\
grossbal0	&Gross loan balance\\
h	&Prediction horizon\\
premod\_ind	&Time indicator: before 2007Q2 financial crisis vs. after\\
qspread	&Spread: note rate – mortgage rate\\
sato2	&Spread at time of origination\\
totpersincyy	&Total personal income year over year growth\\
orig\_fico	&FICO at origination\\
rgdpqq	&Real GDP year over year growth\\
orig\_ltv	&Origination loan to value ratio\\
orig\_cltv	&Origination combined ltv\\
orig\_hpi	&Origination house price index\\
balloon\_in	&Balloon loan indicator\\
homesalesyy	&Home sales year over year growth\\
hpi0	&hpi at snapshot\\
homesalesqq	&Home sales quarter over quarter growth\\
\hline
\end{tabular}
\caption{Homelending Dataset Important Variable Meanings}
\end{center}
\end{table}

In order to make for easier training, we transformed the dataset as follows: 1. we randomly subsampled from non-default cases in order to give the classes equal number of cases, and 2. we applied the transformation $X_{train}'=D(X_{train}-B)$, where $D$ is a diagonal matrix of two divided by the range of each predictor in the training set, and $B$ is a vector of the means of each predictor in the training set. This casts the training set into the subset $[-1,1]^{55}$. We apply the indentical transformation to the test set, $X_{test}'=D(X_{test}-B)$.
	We started with an original network with 3 hidden layers, 10 neurons each. This gave us a trained model using 202 active configurations, leading to a flattened network of 201 neurons. We then pruned the neurons, one-by-one, according to pruneFlatNetwork. For each network, we recorded the test set accuracy and AUC score, which we present below on the following plots. On the x-axes are the flat network sizes, the exception being the rightmost point, which is the original network. On the y-axes are the accuracies and AUC scores, respectively.

\begin{figure}[h!]
\centering
\includegraphics{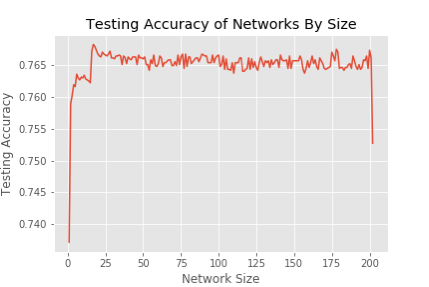}
\includegraphics{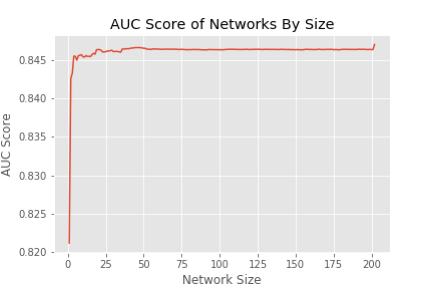}
\caption{Simplified network performances}
\end{figure}
We have identified a collection of sizes as critical points for illustration. We chose the original network, the flattened network, the maximum accuracy achieved (17 neurons), 3 neurons, and 2 neurons. In Table 3, we present more precise values of accuracy and AUC score for these five networks. The results show clearly that we were able to achieve comparable accuracy and AUC scores with much smaller, simpler networks. 

\begin{table}[h!]
\begin{center}
\caption{Network Performance by Size}
\begin{tabular}{ | l || c | c |}
\hline
Network Size & Accuracy & AUC Score\\
\hline
\hline
Original: $[10, 10, 10]$	&0.752721	&0.846993\\
Flattened: $[201]$	&0.766173	&0.846347\\
$[17]$	&0.768289	&0.845846\\
$[3]$	&0.760278	&0.843248\\
$[2]$	&0.758918	&0.842573\\
\hline
\end{tabular}
\end{center}
\end{table}

Our first step in visual interperetation was to extract the network’s configurations, linear equations, and partition. Then, utilizing this information, we use two different main methods for global, visual interperetation and are able to show an explicit and numerical local interperetation as well. First, we make parallel coordinate plots (PC plots) of each predictor’s weight in each linear equation. This allows us to see visually which predictors are most important, along with what influence they have on the decision of the model. We set each linewidth approximately proportional to the number of instances that fall into that configuration for better visualization of region size.  We do this both including all predictors and highlighting only 6 important predictors, according to \cite{axnn}. Second, we make a matrix plot including the same 6 predictors $x_i$ and graph $\text{d} y/\text{d} x_j$  vs. $x_i$, according to \cite{derivatives}, using $\hat{W_j}$ from the linear equation corresponding to the configuration for $x_i$ as $\text{d} y/\text{d} x_j$. This shows us not only the weights given to important predictors, but also key interactions between predictors. Finally, for the 2-neuron network, we include a copy of the two configurations’ linear equations and inequalities defining the regions. All of this information is shown in the Appendix.

In the global interpretation plots, one can see the clear advantage of extremely small networks. On a PC plot, a small network will show fewer lines, which makes it much easier to understand. Also, given few enough regions, with the matrix plot one can also tell exactly which instances tend to fall into what regions globally, in addition to seeing key nonlinearities in great detail. In local interpretation, small, single-hidden-layer networks are very easy to deal with, as they carry with them fewer region boundaries and fewer linear equaitons.

\begin{figure}[h!]
\centering
\caption{PC Plots for highly simplified networks on homelending data}
\subfigure[PC Plot for the 3-neuron simplified network, displaying output dependence on input region-by-region]
{
\includegraphics{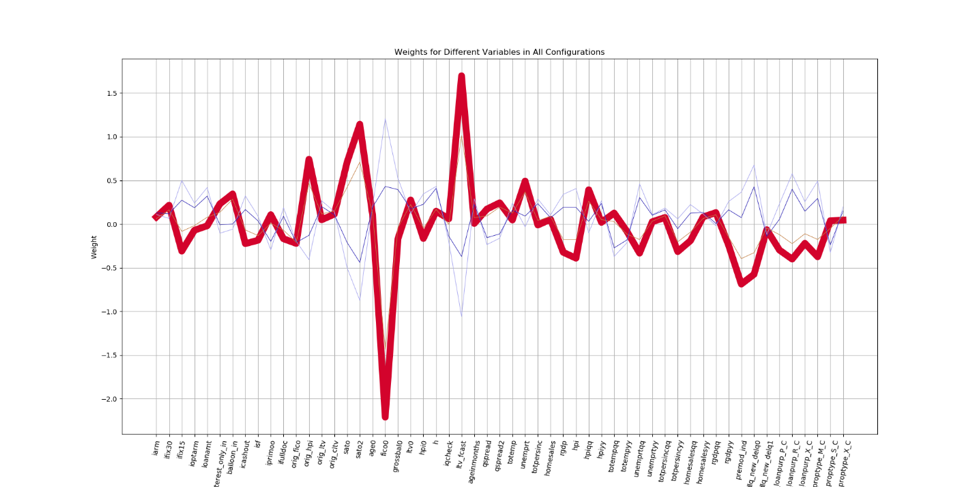}
}

\subfigure[PC Plot for the 2-neuron simplified network--as compared to the above plot, it is clear to see that two pairs of regions were merged and averaged, resulting in a similar model with fewer regions]
{
\includegraphics{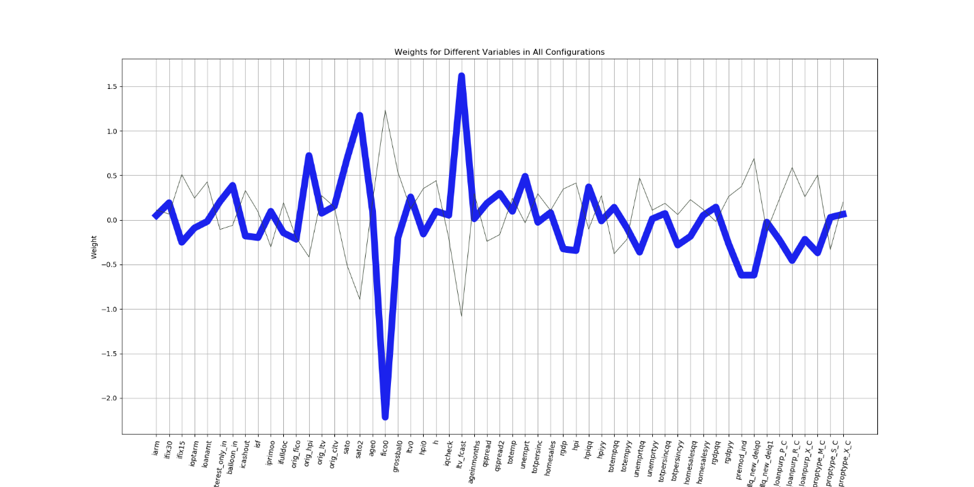}
}
\end{figure}

We include copies of PC plots for the 3-neuron and 2-neuron networks for immediate discussion related to the pruning process. In the first chart, we see the linear equation weights of the 4 active regions in the 3-neuron network. There is one dominant region into which most instances fall, which follows the general pattern of all other networks shown (see Appendix). There are two obvious pairs of regions which follow similar patterns in terms of weight. In the second chart, we see the linear equation weights of the 2 active regions in the 2-neuron network. As before, there is one dominant region, following the general pattern of the other networks.

We bring the reader’s attention to the transition from 4 to 2 regions. As discussed in Section 3, deleting a neuron does in fact merge two pairs of active regions. The method of pruning served to, in a sense, cluster the pairs that are the most similar (i.e. have the highest cosine similarity), returning a network which combines the two pairs of clustered regions into two regions, each a sort of “average” of the corresponding cluster. This is no mistake—in fact, it is exactly the intent of pruning in this way, as described previously in Section 3. 

	Further, we note that this clustering serves to preserve model diversity and significant nonlinearities. This is why, despite the 199 neurons pruned, the 2-neuron network had regions with very different linear equations. We can therefore say that, in a broad sense, the goal of preserving diversity and nonlinearity with a minimal model was achieved. 
	
	One may ask why in the 2-neuron network, the small region has some weights that seem counterintuitive, such as a positive weight for FICO score and a negative weight for LTV forecast. Upon inspection of the matrix plot in Figure 28 , we see that this region is populated only with cases of high FICO score, low LTV forecast, and nondelinquency. These are generally considered to be “safe” loans, unlikely to end in default, and that is reflected in the regional linear equation’s weight of -1.84, a comparatively high bias toward nondefault. Consequently, all of these cases are recognized as nondefault cases by the model. However, when we prune to 1 neuron, the same two regions exist, but the small region becomes a trivial region, and the model becomes less accurate, as the decision boundary must now be parallel to the region boundary. The reader may verify this statement easily\footnote{If $W_1X, B_1, W_2,\text{ and }B_2$ are scalars, then the zero-activation hyperplane of the neuron in $F(X) = \sigma(W_2 \cdot \text{ReLU}(W_1 X + B_1) + B_2)$ leaves $F(X) = \sigma(B_2)$, a constant, and thus either the decision boundary and region boundary are equal or can never cross, both implying that they are parallel. A full proof of this is straightforward.}.
	
	In fact, we see many cases in simplified networks of regions that contain only one predicted class \cite{unwrapping}. Upon removing these lines from our PC plots, we see a much cleaner picture. Figure 29 shows the PC plot of the flat network considering only regions which contain points of multiple classes, and Figure 30 shows the same plot for our maximum accuracy model with 17 neurons. In Figure 31, we consider the four most populated regions and plot only those linear equations.
	
	Additionally, we may consider a simplified network as an additive model of a subset of linear equations from the original network. In this way of thinking, the model output is really---in a heuristic sense---a weighted average of the decisions from each region. We show the 17 linear equations used in our maximum-accuracy model on a PC plot in Figure 33. Although many of these equations have similar slope, there is still enough diversity for the flattened network to recognize important features from differences in these linear equations.

\section{Further Work}

We encourage further investigation in several key areas. Since flattening is a new technique, we would like to better understand its behavior, especially on more datasets. We would like to see how the size of network and size of data affect the performance and advantages provided by flattening. Our results presented here began to show what looks like a correlation between the size of the flattened network and its performance relative to the original network. That is, the larger the original model, the better flattening performs, not only in an absolute sense, but also relative to the original model. We also observed what appears to be a curve that fits flattened network performance as a function of flattened network size, but could not collect enough data to support such a curve of fit. 

One concern that arose during testing is that flattening can produce some linear equations for small regions that appear not to fit the patterns of the original model, but despite this, the flattened model still works well. In our testing, we were able to verify that regions with such strange linear equations were in fact mostly single-instance or single-class regions and led to highly accurate classifications regardless. We would like to investigate this in the future, specifically in a few key areas. First, what tweaks can be made to the algorithm to minimize this phenomenon? For instance, it may be possible to use a different generalized linear model than regularized logistic regression to accomplish our task. Second, are there serious robustness issues to be concerned about, or can we guarantee that such region boundaries with dubious normal vectors will not actively classify significant amounts of data, effectively making these quasi-trivial regions? Third, is there a correlation between the magnitude of our concerns about this and the type of model we are flattening? Finally, is there an effective way to prune or directly change parameters such that these decision boundaries can be corrected?

	Additionally, our methods for simplification raise the question: Is it possible to determine the necessary neurons of a flat network without going through a stepwise pruning process? After flattening, we would like to know which neurons are needed to maintain the necessary nonlinearities. We would also like to know which neurons contribute to the diversity of the model, and which neurons might lead to redundancies and overfitting. Investigation in this area could potentially lead to exciting levels of efficiency for this process and others like it.
	
	Another logical next step is the application of our techniques on nonbinary classification tasks. Our experiments and development focused solely on binary classification, but mathematically speaking, there is no reason why any of the theory included in this paper would not apply to a multiple output network. Therefore, we hope that our techniques will also generalize to nonbinary classification tasks. \em{Note}\em: when performing regression on the output layer in flattening and pruning, nonbinary classification would require performing regression for each class.

	Finally, we would like to investigate these reduction techniques applied to regression tasks. With a regression objective, the network simply does not use the sigmoid activation on the output layer, which makes the task of flattening more difficult. Using sigmoid activation allows us only to fit the zero point of the network’s linear equations. However, with regression tasks, we must also be able to fit the correct slope. Some exploration of this has already been done in \cite{unwrapping}, and the initial results are promising. Our hope is that despite this added difficulty, these techniques will nevertheless result in good performance.
	
\section{Conclusions}

In this paper, we investigated the working mechanisms of Piecewise Linear Neural Networks as piecewise linear functions, applying local linear models within learned regions. We introduced machinery toward simplification of over-parametrized deep networks, along with single-hidden-layer representations, without sacrificing model performance. Our approach, when applied to Piecewise Linear Neural Networks, produces well-performing and more interpretable models.

	We exploit the piecewise-linear activation function to derive a network’s linear equations. As those equations are naturally pertinent to the classification task presented to us, we utilize them in the creation of a flat network, where our novel hidden-layer transformation provides the coordinates of an instance with respect to each linear equation. We then solve the convex optimization problem of logistic regression to determine the output layer parameters. This process, called “flattening,” was tested on both simulated and real data, and was robust for sufficiently large networks.
	
	After designing a flattened network, we pruned neurons stepwise using L2-penalized logistic regression. This comes on the basis of intuition regarding the geometric orientation of neighboring regions’ decision boundaries. We saw in every case that there is a significant amount that we may prune a flattened network freely without loss in performance. Such a small model is easy to explain and results in much faster computation for real-time use. In practice, we were able to simplify an over-parameterized model into one of minimal size with equal testing accuracy and interperet it, both globally and locally. Our analysis has illustrated the potential of these methods for study and application in model simplification.
\pagebreak
\bibliography{ssmbib}{}
\bibliographystyle{IEEEtran}

\appendix

\section{Figures}

\begin{figure}[h]
\includegraphics{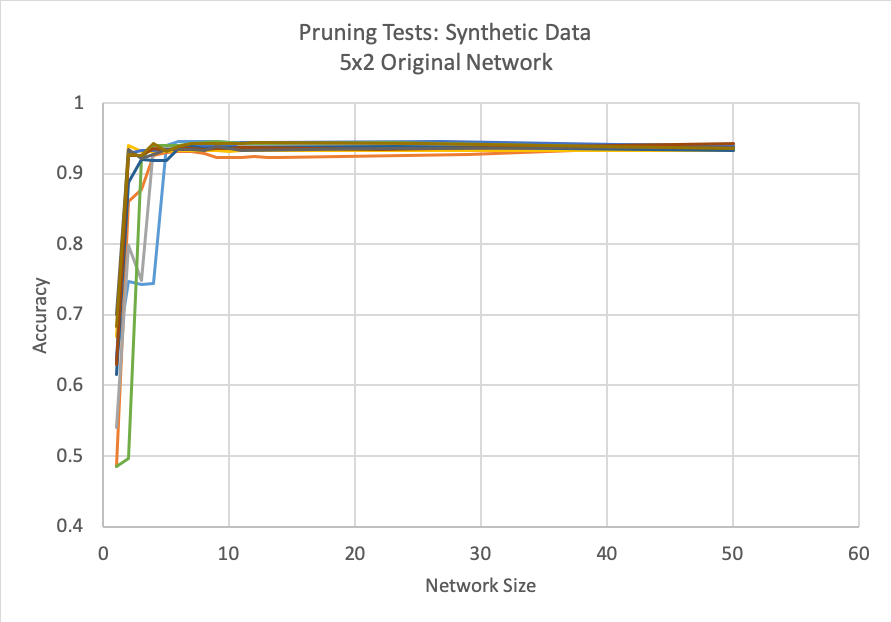}
\caption{}
\end{figure}

\begin{figure}
\includegraphics{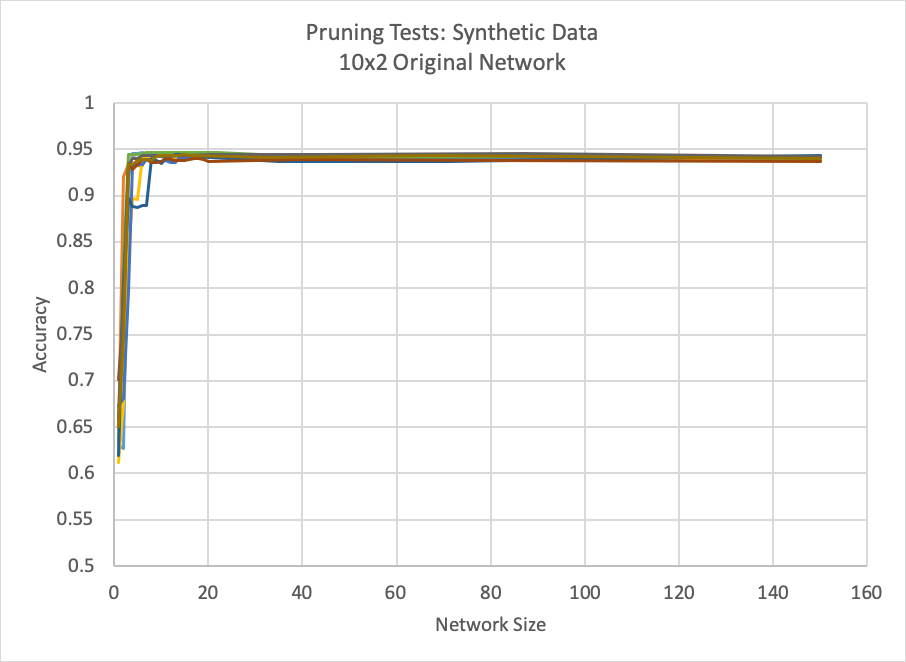}
\caption{}
\end{figure}

\begin{figure}
\includegraphics{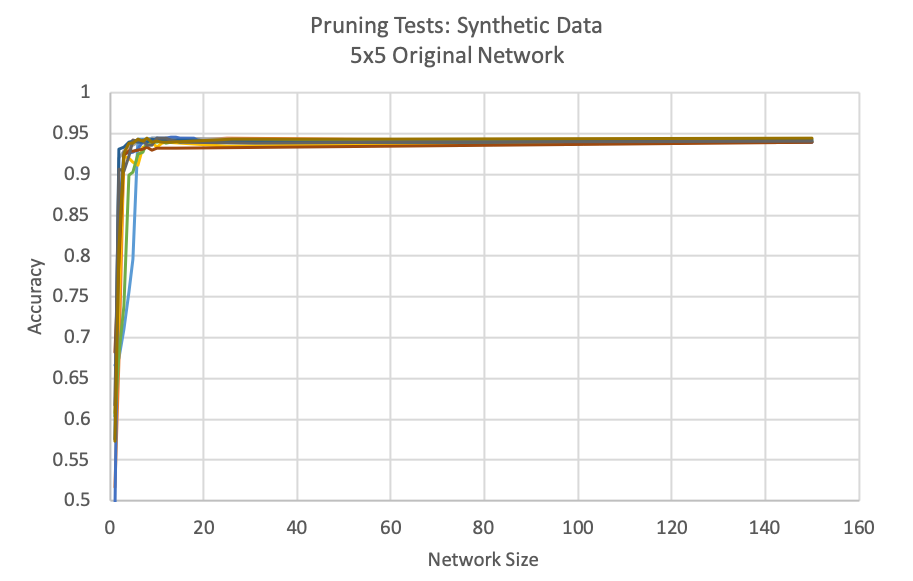}
\renewcommand\thefigure{\ref{figure:fig11}}
\caption{}
\end{figure}
\addtocounter{figure}{-1}

\begin{figure}
\includegraphics{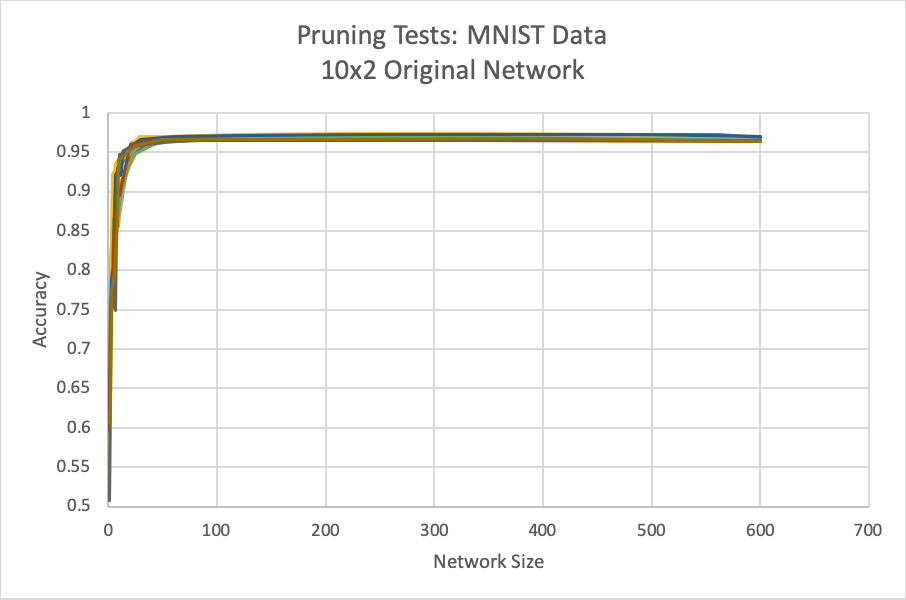}
\caption{}
\end{figure}

\begin{figure}
\includegraphics{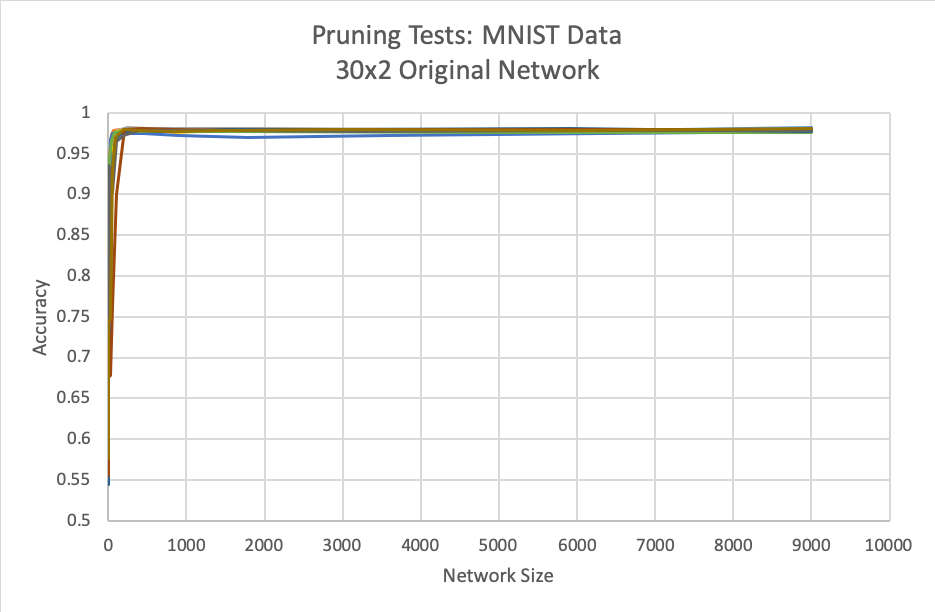}
\caption{}
\end{figure}

\begin{figure}
\includegraphics{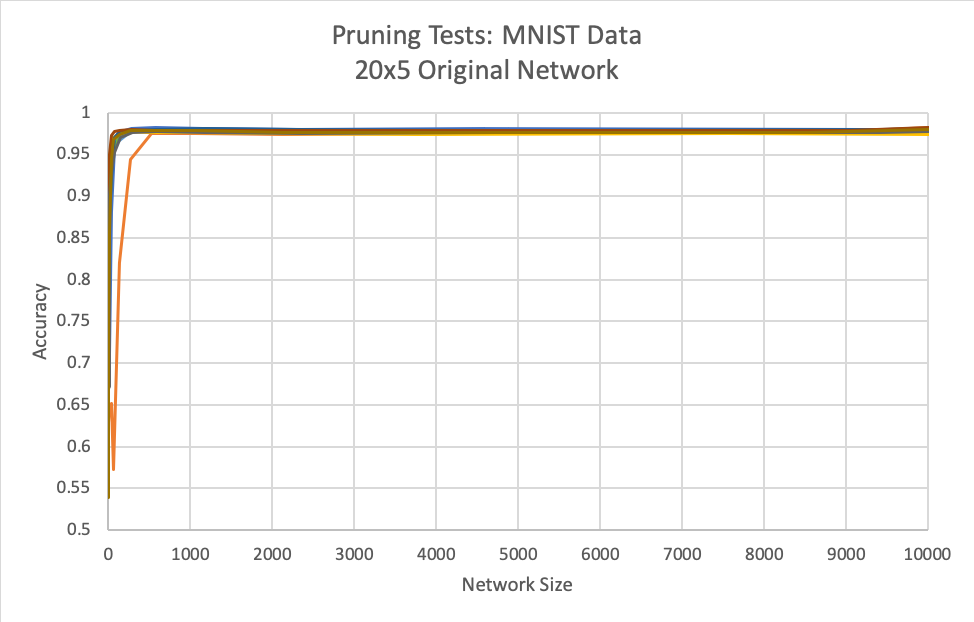}
\caption{}
\end{figure}

\begin{figure}
\centering
\includegraphics{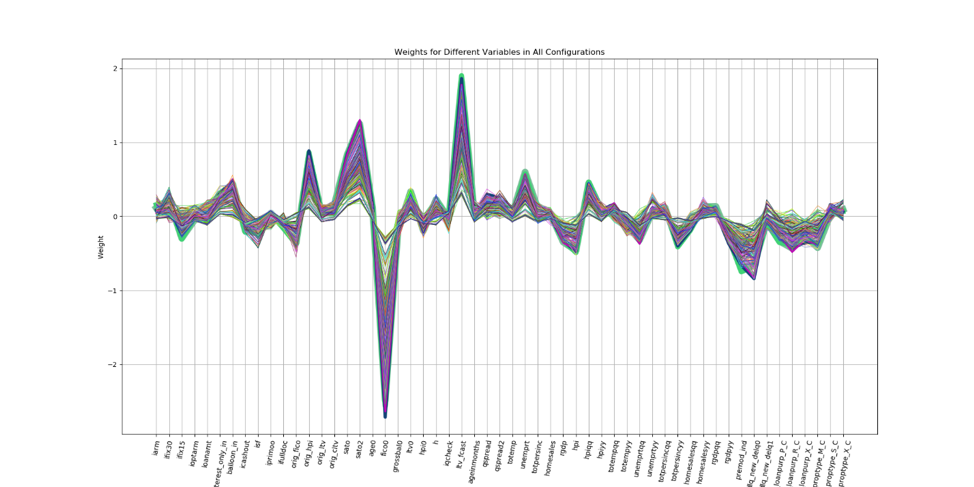}
\caption{PC plot, all predictors, original network}
\end{figure}

\begin{figure}
\centering
\includegraphics{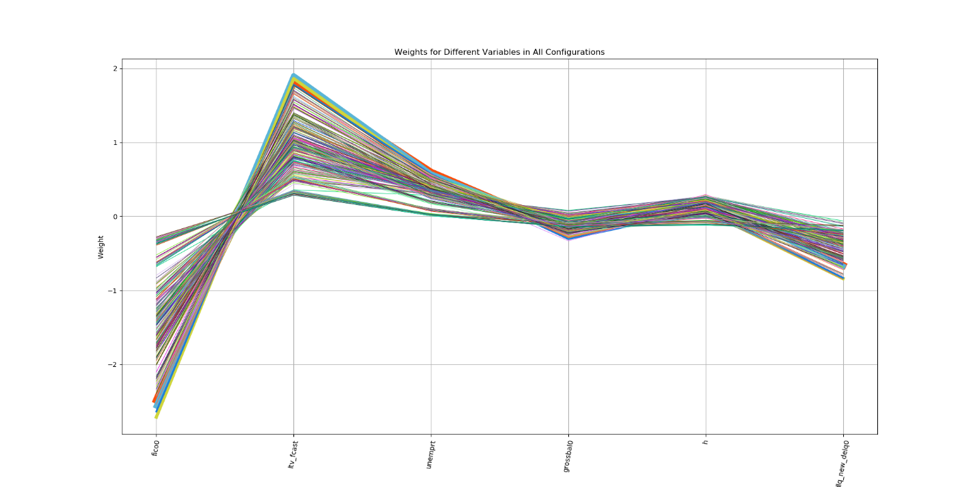}
\caption{PC plot, main predictors, original network}
\end{figure}

\begin{figure}
\centering
\includegraphics{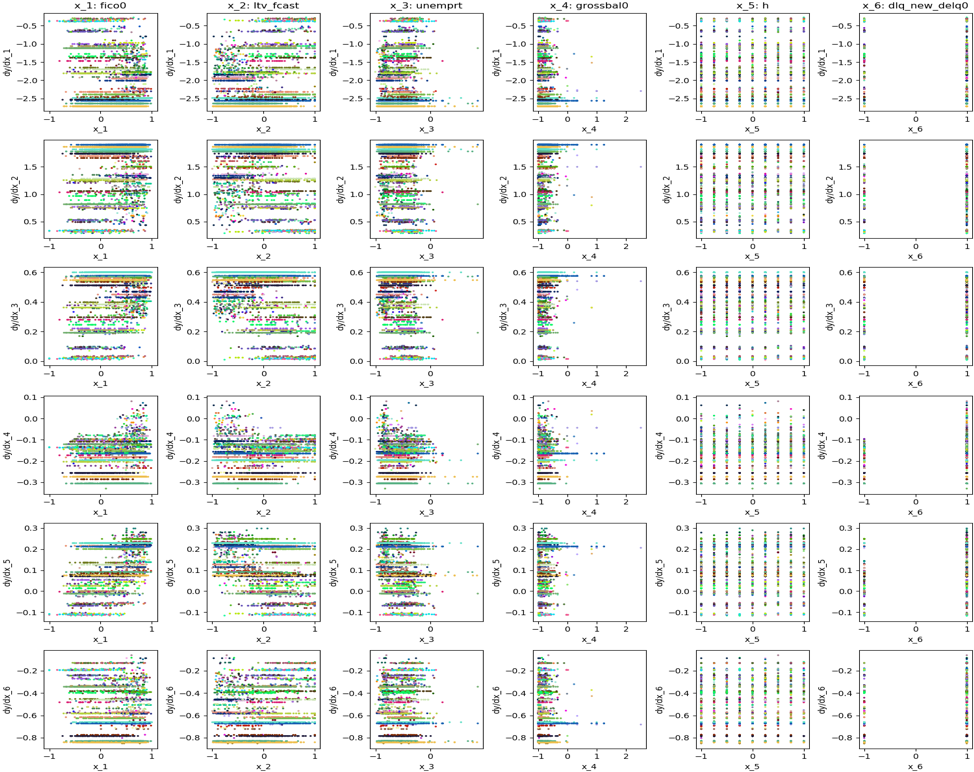}
\caption{Matrix plot, original network}
\end{figure}

\begin{figure}
\centering
\includegraphics{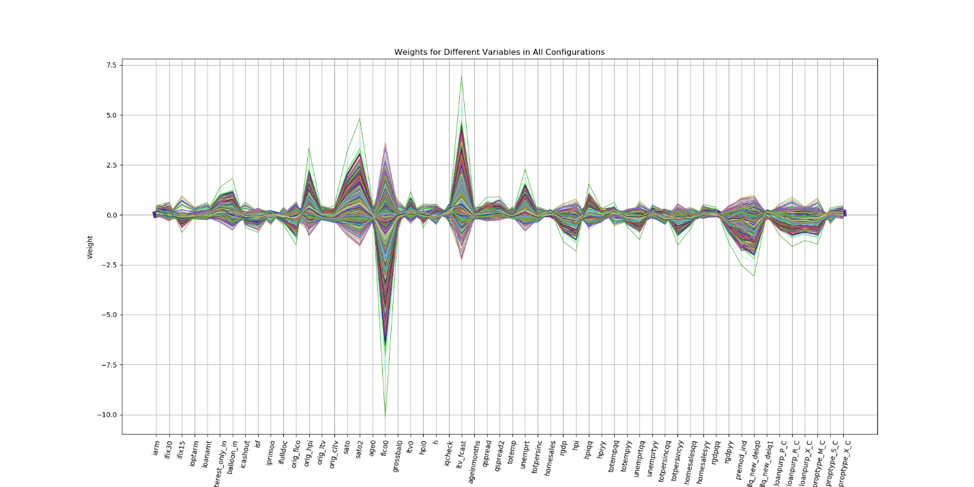}
\caption{PC plot, all predictors, flattened network}
\end{figure}

\begin{figure}
\centering
\includegraphics{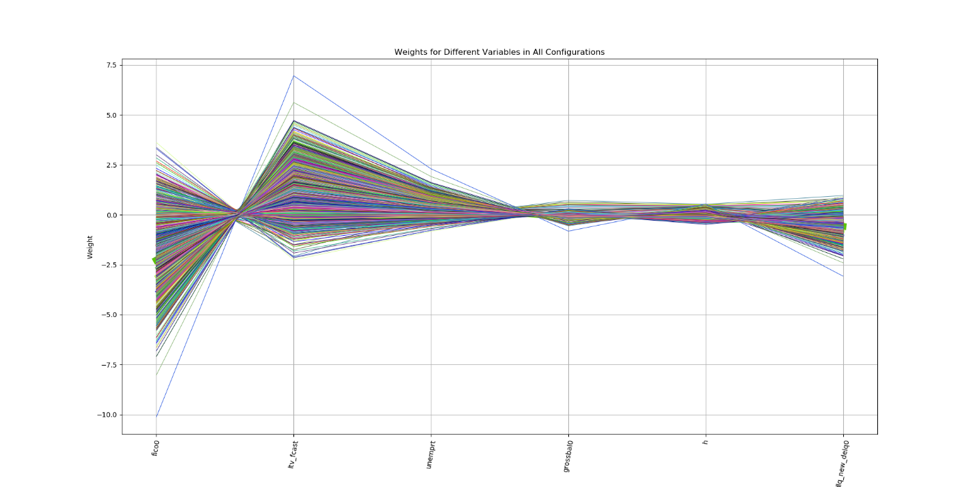}
\caption{PC plot, main predictors, flattened network}
\end{figure}

\begin{figure}
\centering
\includegraphics{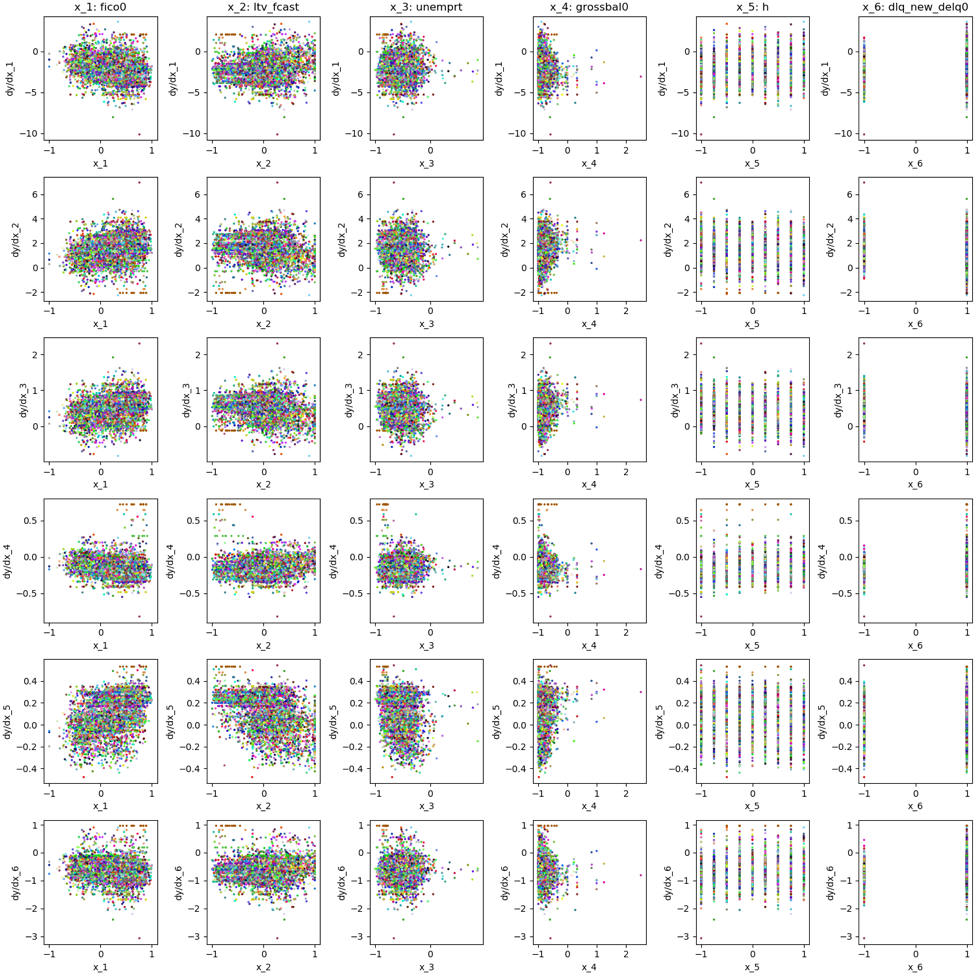}
\caption{Matrix plot, flattened network}
\end{figure}

\begin{figure}
\centering
\includegraphics{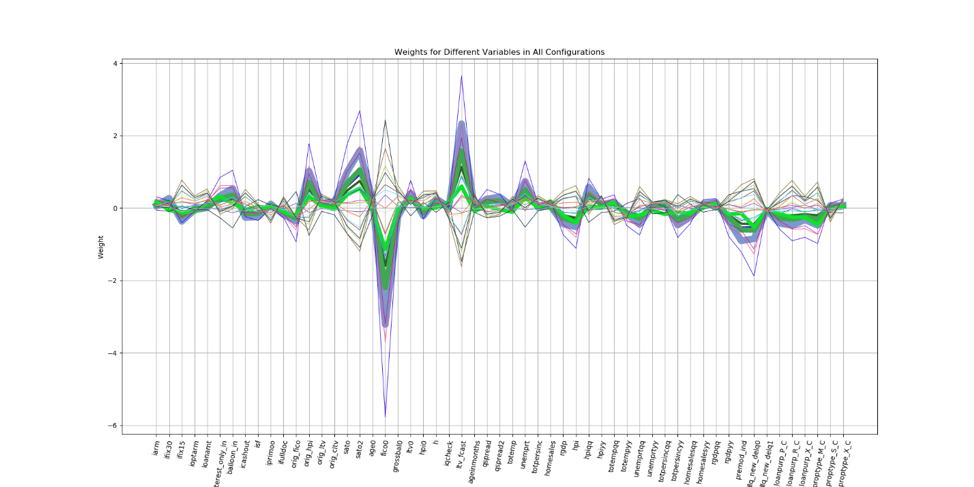}
\caption{PC plot, 17 neurons, all predictors}
\end{figure}

\begin{figure}
\centering
\includegraphics{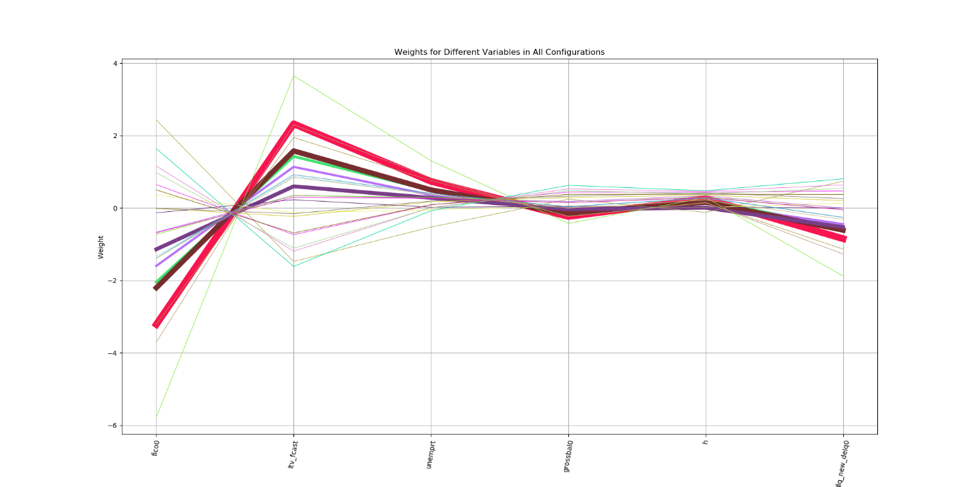}
\caption{PC plot, 17 neurons, main predictors}
\end{figure}

\begin{figure}
\centering
\includegraphics{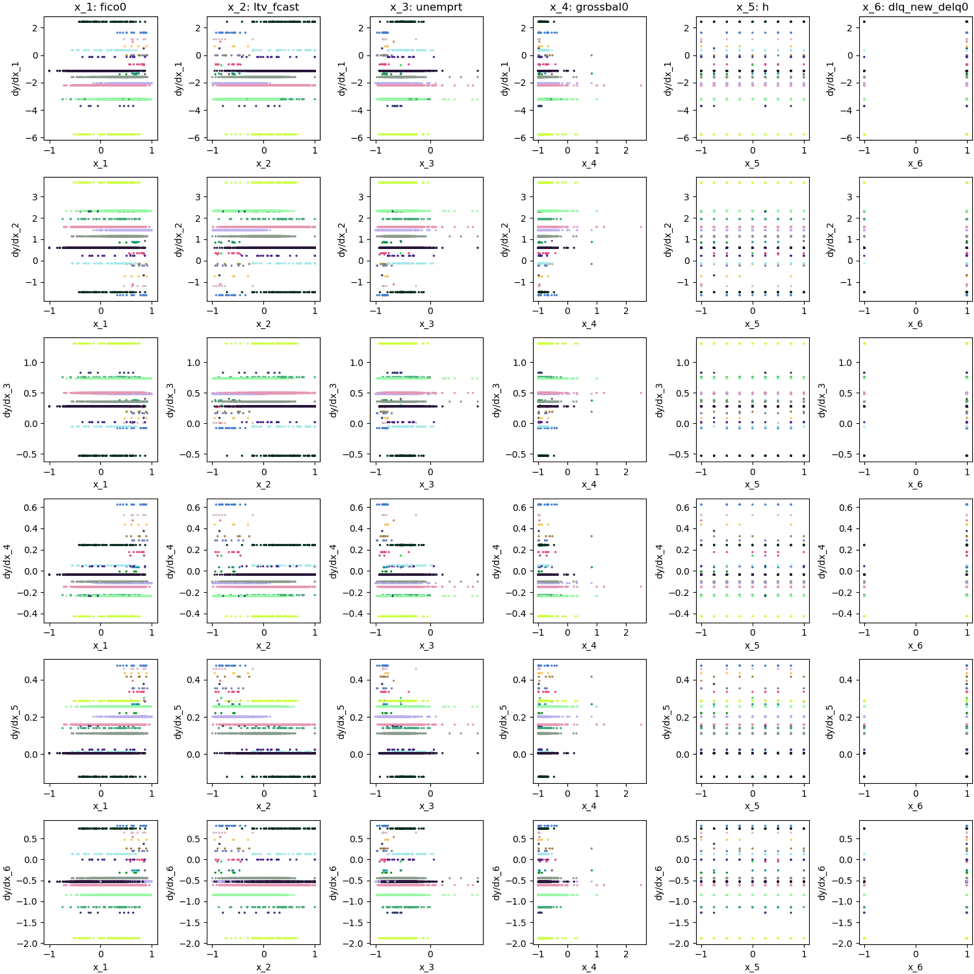}
\caption{Matrix plot, 17 neurons}
\end{figure}

\begin{figure}
\centering
\includegraphics{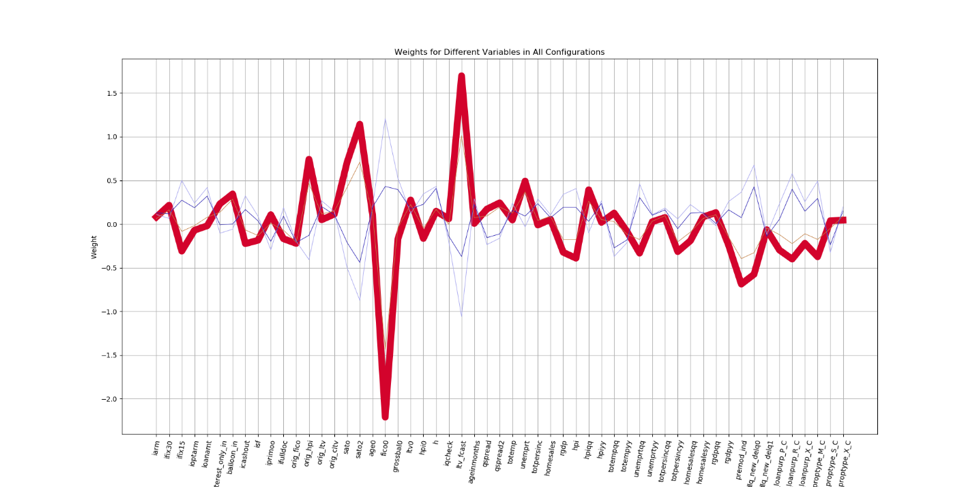}
\caption{PC plot, 3 neurons, all predictors}
\end{figure}

\begin{figure}
\centering
\includegraphics{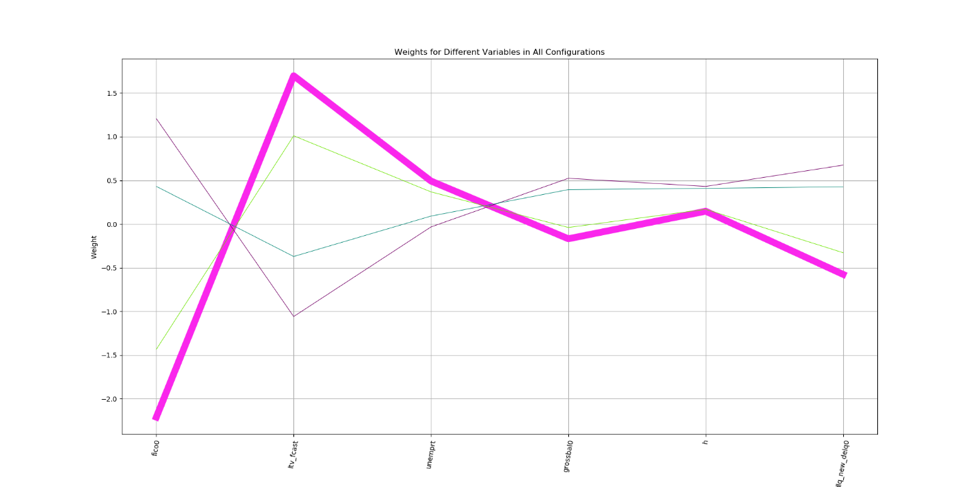}
\caption{PC plot, 3 neurons, main predictors}
\end{figure}

\begin{figure}
\centering
\includegraphics{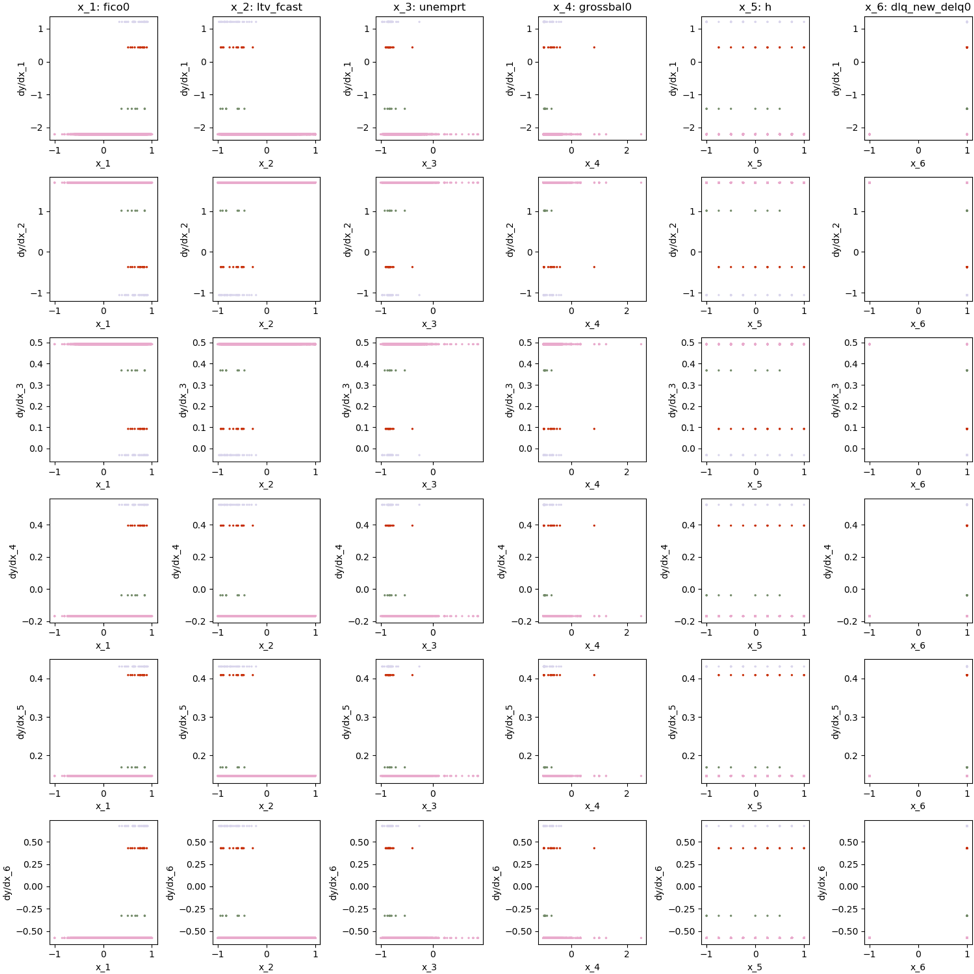}
\caption{Matrix plot, 3 neurons}
\end{figure}

\begin{figure}
\centering
\includegraphics{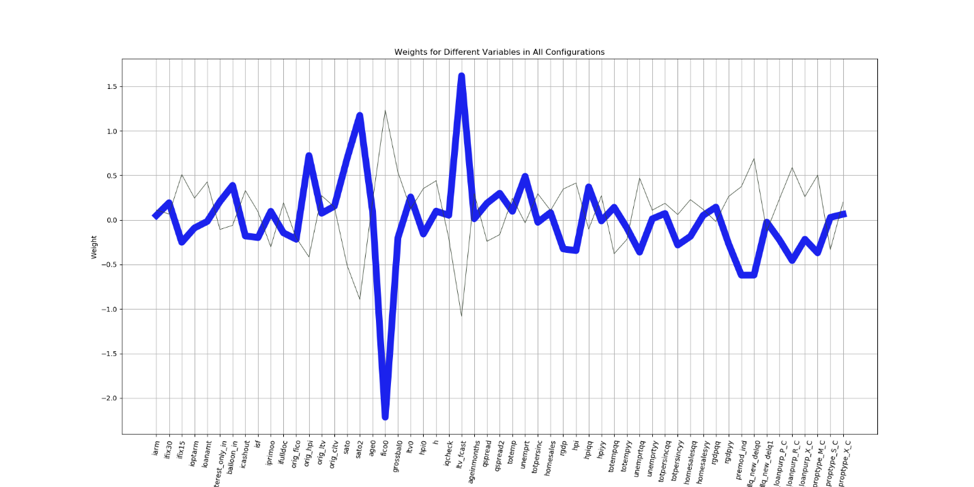}
\caption{PC plot, 2 neurons, all predictors}
\end{figure}

\begin{figure}
\centering
\includegraphics{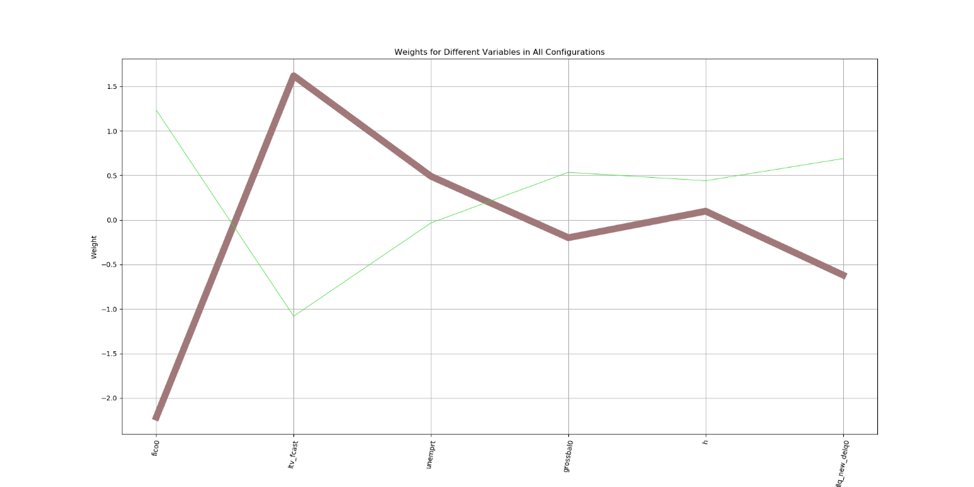}
\caption{PC plot, 2 neurons, main predictors}
\end{figure}

\begin{figure}
\centering
\includegraphics{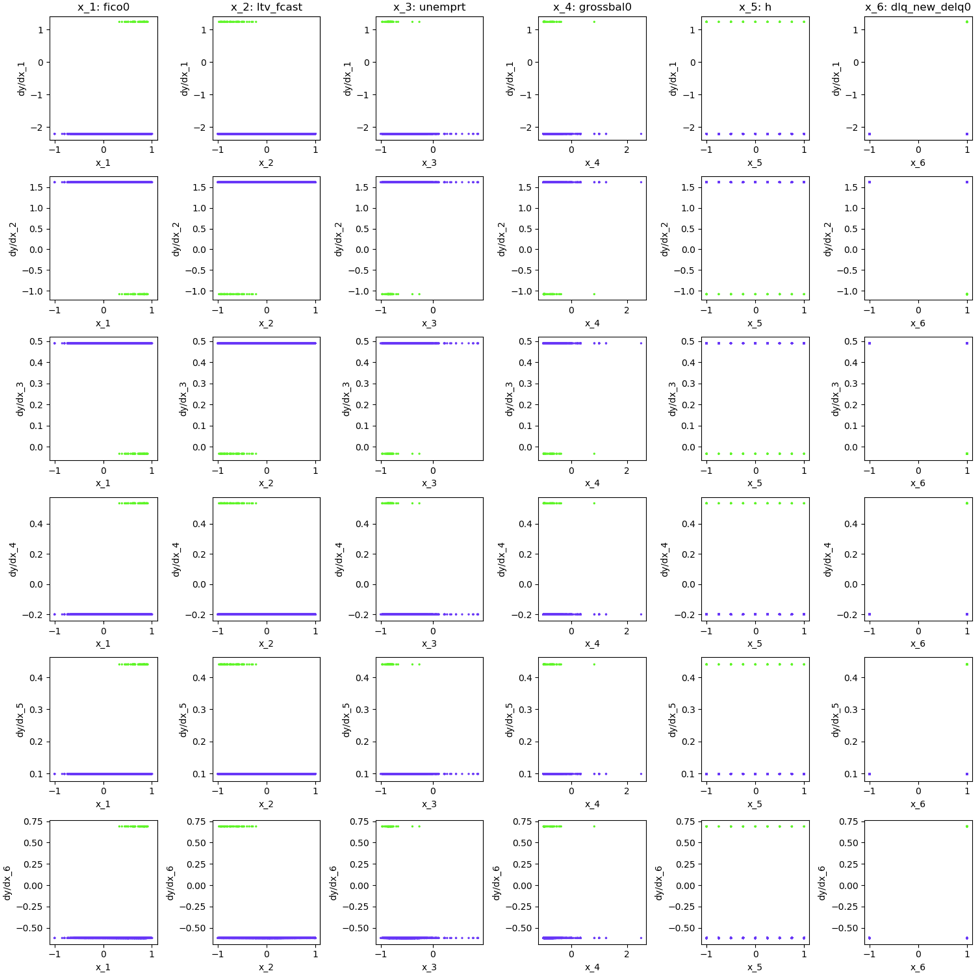}
\caption{Matrix plot, 2 neurons}
\end{figure}

\begin{figure}
\centering
\includegraphics{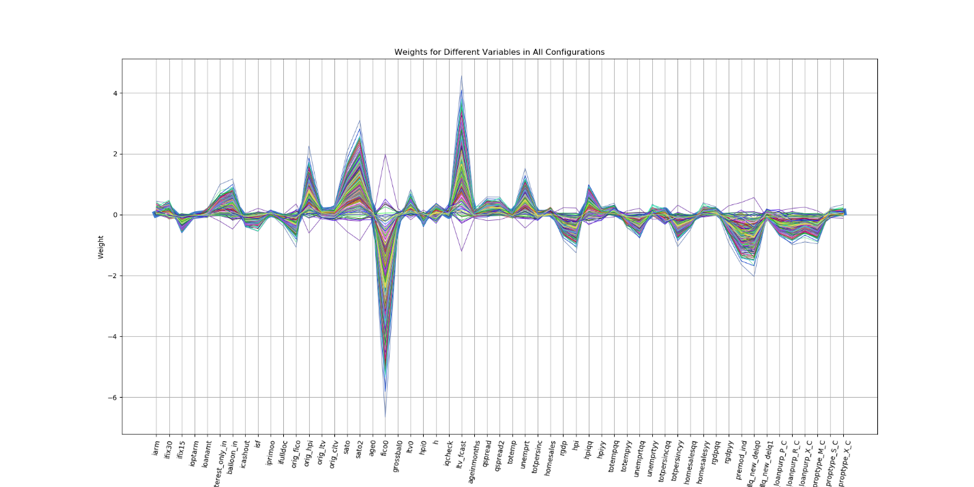}
\caption{Flattened network linear equations, mixed classification regions only}
\end{figure}

\begin{figure}
\centering
\includegraphics{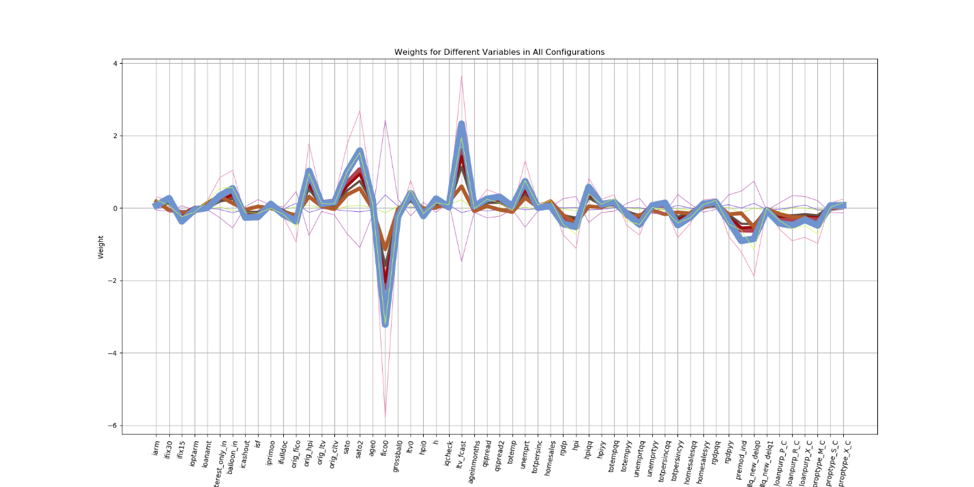}
\caption{17 neuron network linear equations, mixed classification regions only}
\end{figure}

\begin{figure}
\centering
\includegraphics{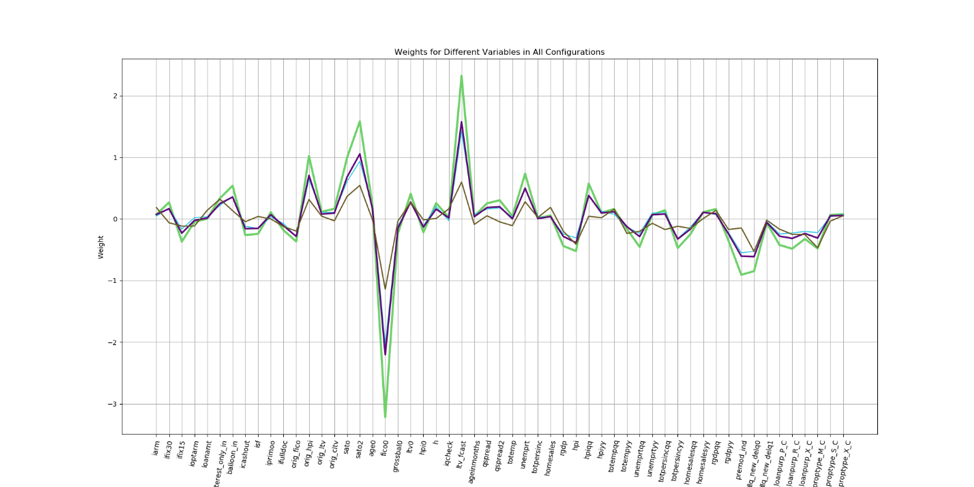}
\caption{17 neuron network linear equations for the 4 most populated regions}
\end{figure}

\begin{figure}
\centering
\includegraphics{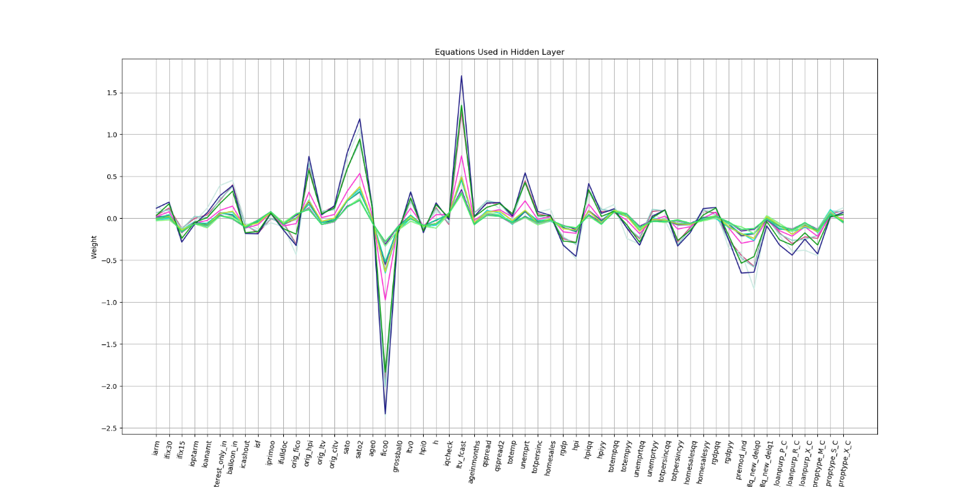}
\caption{Original network linear equations used in the hidden layer of the 17 neuron network}

\end{figure}
\pagebreak
\section{Exact Interpretation of 2-Neuron Network}

\subsection{Region Boundary Inequalities}

Configuration '[1.0, 1.0]':\\
\noindent\makebox[\linewidth]{\rule{\linewidth}{0.4pt}}

 [-0.01137126,0.02314699,-0.13846233,-0.06100354,-0.08073598,
 
           0.05709722,0.0815693 ,-0.09272958,-0.0524761 ,0.07260806,
           
          -0.06113361,-0.00265612,0.20699275,-0.03590856,0.00161831,
          
           0.21816143,0.3759866 ,-0.01966394,-0.6272964 ,-0.13365561,
           
           0.0242728 ,-0.09293875,-0.06223787,0.04732117,0.49132606,
           
          -0.05144775,0.0777434 ,0.08458227,-0.02664522,0.09520718,
          
          -0.05818958,-0.00427326,-0.12243342,-0.13807969,0.08617731,
          
          -0.05131209,0.09480578,0.02372852,-0.1513769 ,-0.0176851 ,
          
          -0.02072435,-0.06232166,-0.07501411,-0.01217248,0.02959712,
          
          -0.09653986,-0.18066609,-0.23805696,0.01819998,-0.08371982,
          
          -0.18994443,-0.08700642,-0.15868974,0.06582268,-0.02477659] $\cdot$ x+0.2992687523365021 $\geq$ 0 \\
\noindent\makebox[\linewidth]{\rule{\linewidth}{0.4pt}}

   [-0.0297007 ,-0.01746409,-0.1348328 ,-0.06489145,-0.11284263,
   
           0.02821928,0.01590886,-0.0872122 ,-0.02404664,0.08001817,
           
          -0.05047452,0.05327386,0.10967118,-0.07208531,-0.03855323,
          
           0.13465449,0.23582901,-0.05510059,-0.32595938,-0.1414809 ,
           
          -0.032949  ,-0.09323773,-0.11643595,0.05474446,0.2857164 ,
          
          -0.07842055,0.06324827,0.04362519,-0.06439807,0.00879428,
          
          -0.07768828,-0.02811446,-0.09182094,-0.10966326,0.02714658,
          
          -0.07180463,0.10020324,0.0579771 ,-0.12438878,-0.02898008,
          
          -0.04939212,-0.01613373,-0.06033386,-0.03120083,0.00489322,
          
          -0.06965625,-0.09872035,-0.18221949,0.03289123,-0.0625117 ,
          
          -0.15544024,-0.06904704,-0.1329765 ,0.08810747,-0.05295335] $\cdot$ x+ 0.3402073085308075 $\geq$ 0 \\
\noindent\makebox[\linewidth]{\rule{\linewidth}{0.4pt}}
\noindent\makebox[\linewidth]{\rule{\linewidth}{0.4pt}}\\
\\
Configuration ‘[0.0, 1.0]’: \\
\noindent\makebox[\linewidth]{\rule{\linewidth}{0.4pt}}
[-0.01137126,0.02314699,-0.13846233,-0.06100354,-0.08073598,

           0.05709722,0.0815693 ,-0.09272958,-0.0524761 ,0.07260806,
           
          -0.06113361,-0.00265612,0.20699275,-0.03590856,0.00161831,
          
           0.21816143,0.3759866 ,-0.01966394,-0.6272964 ,-0.13365561,
           
           0.0242728 ,-0.09293875,-0.06223787,0.04732117,0.49132606,
           
          -0.05144775,0.0777434 ,0.08458227,-0.02664522,0.09520718,
          
          -0.05818958,-0.00427326,-0.12243342,-0.13807969,0.08617731,
          
          -0.05131209,0.09480578,0.02372852,-0.1513769 ,-0.0176851 ,
          
          -0.02072435,-0.06232166,-0.07501411,-0.01217248,0.02959712,
          
          -0.09653986,-0.18066609,-0.23805696,0.01819998,-0.08371982,
          
          -0.18994443,-0.08700642,-0.15868974,0.06582268,-0.02477659] $\cdot$ x+0.2992687523365021 $\leq$ 0\\
\noindent\makebox[\linewidth]{\rule{\linewidth}{0.4pt}}

   [-0.0297007 ,-0.01746409,-0.1348328 ,-0.06489145,-0.11284263,
   
           0.02821928,0.01590886,-0.0872122 ,-0.02404664,0.08001817,
           
          -0.05047452,0.05327386,0.10967118,-0.07208531,-0.03855323,
          
           0.13465449,0.23582901,-0.05510059,-0.32595938,-0.1414809 ,
           
          -0.032949  ,-0.09323773,-0.11643595,0.05474446,0.2857164 ,
          
          -0.07842055,0.06324827,0.04362519,-0.06439807,0.00879428,
          
          -0.07768828,-0.02811446,-0.09182094,-0.10966326,0.02714658,
          
          -0.07180463,0.10020324,0.0579771 ,-0.12438878,-0.02898008,
          
          -0.04939212,-0.01613373,-0.06033386,-0.03120083,0.00489322,

          -0.06965625,-0.09872035,-0.18221949,0.03289123,-0.0625117 ,
          
          -0.15544024,-0.06904704,-0.1329765 ,0.08810747,-0.05295335] $\cdot$ x+ 0.3402073085308075 $\geq$ 0 \\
\noindent\makebox[\linewidth]{\rule{\linewidth}{0.4pt}}
\subsection{Local Linear Equations}

Configuration '[1.0, 1.0]': \\
\noindent\makebox[\linewidth]{\rule{\linewidth}{0.4pt}}
z=[ 0.04985788,  0.19329724, -0.25098094, -0.08981898, -0.01687544,  0.20708205,
   0.3881659,  -0.17976153, -0.19746734,  0.09637803, -0.14507459, -0.21613982,
   0.7228515,   0.07533223,  0.15474583,  0.68972653,  1.1744691,   0.1003669,
  -2.2148368,  -0.19941007,  0.2580663,  -0.15811606,  0.09839442,  0.05300003,
   1.6197107,   0.013887,    0.18804693,  0.29987326,  0.09716716,  0.49004254,
  -0.02594015,  0.08287157, -0.32559517, -0.34409657,  0.37098074, -0.01039606,
   0.1420271,  -0.08890787, -0.36147746,  0.01242725,  0.07294263, -0.28151938,
  -0.18407075,  0.05112905,  0.14417088, -0.26712084, -0.6195735,  -0.6191404,
  -0.02439348, -0.22368321, -0.45599556, -0.21702437, -0.3691847,   0.02847909,
   0.06414174]  $\cdot$ x -0.19860479 \\
\noindent\makebox[\linewidth]{\rule{\linewidth}{0.4pt}}
\noindent\makebox[\linewidth]{\rule{\linewidth}{0.4pt}}\\
\\
Configuration '[0.0, 1.0]': \\
\noindent\makebox[\linewidth]{\rule{\linewidth}{0.4pt}}
z = [ 0.11236069,  0.06606839,  0.5100859,   0.24549079,  0.42689487, -0.10675634,
  -0.0601848,   0.3299324,   0.09097083, -0.30271667,  0.19095011, -0.2015403,
  -0.41489697,  0.27270588,  0.14585067, -0.5094113,  -0.8921645,   0.20845099,
   1.2331367,   0.53523624,  0.12464934,  0.35272756,  0.44048873, -0.20710373,
  -1.0808935,   0.2966727,  -0.23927447, -0.16503844,  0.2436243,  -0.03326962,
   0.2939025,   0.10635981,  0.34736773,  0.414867,   -0.1026982,   0.27164403,
  -0.37907878, -0.21933313,  0.47057506,  0.1096345,   0.18685529,  0.06103549,
   0.22824897,  0.11803584, -0.01851154,  0.2635165,   0.37346888,  0.6893544,
  -0.12443078,  0.23648798,  0.58804584,  0.2612118,   0.5030633,  -0.3333193,
   0.20032777] $\cdot$ x - 1.84355396\\
\noindent\makebox[\linewidth]{\rule{\linewidth}{0.4pt}}

\end{document}